\documentclass[letterpaper]{article} 
\usepackage{aaai2026}  
\usepackage{times}  
\usepackage{helvet}  
\usepackage{courier}  
\usepackage[hyphens]{url}  
\usepackage{pifont}
\usepackage{amsmath}
\usepackage{graphicx} 
\usepackage{natbib}  
\usepackage{caption} 
\usepackage{amssymb}
\usepackage{algorithm}
\usepackage{algorithmic}
\usepackage{subcaption}
\usepackage{booktabs}  
\usepackage[table]{xcolor}  
\usepackage{multirow}  
\usepackage{array}  
\usepackage{newfloat}
\usepackage{listings}
\DeclareCaptionStyle{ruled}{labelfont=normalfont,labelsep=colon,strut=off} 
\floatstyle{ruled}
\newfloat{listing}{tb}{lst}{}
\floatname{listing}{Listing}
\title{H-GAR: A Hierarchical Interaction Framework via Goal-Driven \\ Observation-Action Refinement for Robotic Manipulation}
\author{
    Yijie Zhu\textsuperscript{\rm 1,2},~
    Rui Shao\textsuperscript{\rm 1,3}\thanks{Corresponding authors.},~
    Ziyang Liu\textsuperscript{\rm 1},~
    Jie He\textsuperscript{\rm 1},~
    Jizhihui Liu\textsuperscript{\rm 1},~
    Jiuru Wang\textsuperscript{\rm 4},~
    Zitong Yu\textsuperscript{\rm 2,5}\footnotemark[1]
}
\affiliations{
    \textsuperscript{\rm 1} Harbin Institute of Technology, Shenzhen\quad 
    \textsuperscript{\rm 2} Great Bay University\quad 
    \textsuperscript{\rm 3} Shenzhen Loop Area Institute\\
     \textsuperscript{\rm 4} Linyi University\quad 
    \textsuperscript{\rm 5} Dongguan Key Laboratory for Intelligence and Information Technology\\
    zyj99\_hitsz@stu.hit.edu.cn\quad shaorui@hit.edu.cn\quad yuzitong@gbu.edu.cn\\
    \textcolor[HTML]{0D47A1}{\url{https://github.com/JiuTian-VL/H-GAR}}
}

\usepackage{bibentry}

\begin{document}

\maketitle

\begin{abstract}
Unified video and action prediction models hold great potential for robotic manipulation, as future observations offer contextual cues for planning, while actions reveal how interactions shape the environment. However, most existing approaches treat observation and action generation in a monolithic and goal-agnostic manner, often leading to semantically misaligned predictions and incoherent behaviors.
To this end, we propose \textbf{H-GAR}, a \textbf{H}ierarchical interaction framework via \textbf{G}oal-driven observation-\textbf{A}ction \textbf{R}efinement.
To anchor prediction to the task objective, H-GAR first produces a goal observation and a coarse action sketch that outline a high-level route toward the goal. To enable explicit interaction between observation
and action under the guidance of the goal observation for more coherent decision-making, we devise two synergistic modules. \textbf{(1) Goal-Conditioned Observation Synthesizer (GOS)} synthesizes intermediate observations based on the coarse-grained actions and the predicted goal observation. \textbf{(2) Interaction-Aware Action Refiner (IAAR)} refines coarse actions into fine-grained, goal-consistent actions by leveraging feedback from the intermediate observations and a \textbf{Historical Action Memory Bank} that encodes prior actions to ensure temporal consistency. By integrating goal grounding with explicit action-observation interaction in a coarse-to-fine manner, H-GAR enables more accurate manipulation. Extensive experiments on both simulation and real-world robotic manipulation tasks demonstrate that H-GAR achieves state-of-the-art performance. 

\end{abstract}


\section{1~~Introduction}
\begin{figure}[t]
    \centering
    \includegraphics[width=1\linewidth]{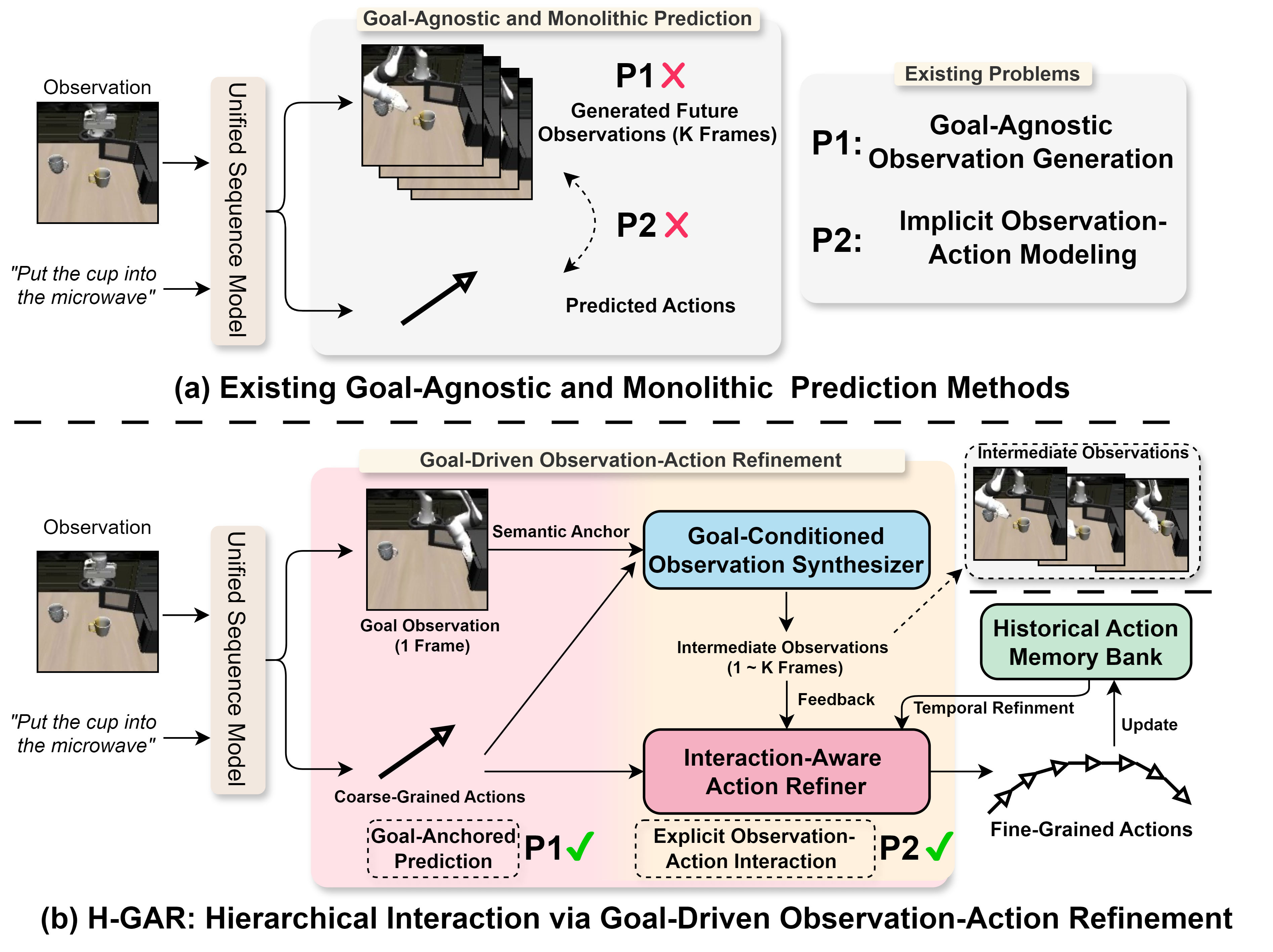}
    \vspace{-17pt}
    \caption{\textbf{Comparison Between Existing Methods and Our Proposed H-GAR}. \textbf{(a)} Existing approaches follow a goal-agnostic and monolithic prediction paradigm, jointly generating observations and actions without explicit goal grounding or structured interaction.
\textbf{(b)} H-GAR introduces a goal-conditioned observation synthesizer and an interaction-aware action refiner, enabling goal-anchored prediction and explicit observation-action interaction.}
\vspace{-1pt}
    \label{fig:intro}
\end{figure}
Effective planning and manipulation in robotics~\cite{chatvla, cogact, dexgraspvla, autoeval, gr1, roboagent, zhu2025emosym, zhu2025uniemo, li2025lion, xie2025gui, shao2024detecting, li2025cogvla} require the ability to anticipate both how the environment evolves and how actions unfold over time. To this end, a growing body of research~\cite{cot, liang2024dreamitate, li2025unified} has shown that jointly predicting future visual observations and action sequences brings substantial benefits to robotic manipulation. This owes to the fact that anticipated visual observations provide rich contextual cues, including spatial layouts, object affordances, and interaction dynamics, which help guide action planning. In parallel, predicted actions expose the causal structure of interactions, enabling more accurate forecasting of how scenes evolve over time. This bidirectional coupling between vision and action has proven critical for modeling real-world dynamics and achieving coherent manipulation.

However, most existing approaches~\cite{li2025unified, du2023learning, zhu2025unified, zhang2025up, cot} adopt a monolithic and goal-agnostic generation strategy. As illustrated in Fig.~\ref{fig:intro}(a), they directly predict the entire rollout of future observations and actions from the current observation and instruction, without explicitly modeling how actions influence observations, or how the target goal should guide the trajectory. This design leads to two fundamental limitations: \textbf{(1) Goal-Agnostic Observation Generation.} Without an explicit goal, models lack semantic guidance during rollout prediction. This often results in visually plausible but task-irrelevant sequences, undermining interpretability and degrading downstream planning accuracy~\cite{liu2025efficient, hu2023planning, brohan2023can}.
\textbf{(2) Implicit Observation-Action Modeling.} Observations and actions are often generated in parallel or via loosely coupled pathways, lacking explicit modeling of their causal interplay. This weakens temporal coherence, limits adaptability, and hinders the mutual reinforcement between perception and manipulation essential for decision-making.


To address these limitations, we propose \textbf{H-GAR}, a \textbf{H}ierarchical interaction framework via \textbf{G}oal-driven observation-\textbf{A}ction \textbf{R}efinement. It introduces explicit goal grounding and structured bidirectional interaction between observation and action. Concretely, as shown in Fig.~\ref{fig:intro}(b), to provide a semantically grounded plan, H-GAR first predicts a goal observation that represents the final visual state to be achieved, alongside a sequence of coarse-grained actions leading toward the goal.
Subsequently, to enable explicit bidirectional interaction between observation and action under the guidance of the goal observation for more coherent decision-making, H-GAR employs two key modules: \textbf{(1) Goal-Conditioned Observation Synthesizer (GOS).} To provide semantically meaningful visual guidance aligned with the task goal, GOS synthesizes intermediate observations conditioned on the predicted goal observation and the initially generated coarse-grained actions. These synthesized observations serve as goal-consistent visual context that bridges the gap between high-level task intent and low-level execution. \textbf{(2) Interaction-Aware Action Refiner (IAAR).} To refine the initial coarse-grained actions into fine-grained, temporally coherent commands, IAAR first leverages a \textbf{Historical Action Memory Bank}, which encodes prior fine-grained actions to guide behaviorally consistent and temporally grounded refinements. It then integrates goal-aligned visual feedback from the intermediate observations synthesized by GOS to further adjust and refine each coarse action.
Through this hierarchical coarse-to-fine and goal-conditioned design, H-GAR effectively addresses the limitations of existing approaches. The predicted goal observation anchors generation to the task objective, mitigating semantic drift. while the explicit bidirectional interaction between GOS and IAAR overcomes the implicit modeling of observation and action. By refining coarse action plans into temporally coherent and physically plausible actions, H-GAR enables adaptive, semantically aligned planning, surpassing prior monolithic and goal-agnostic methods in robotic manipulation tasks.

We conduct comprehensive evaluations of H-GAR on both simulation benchmark and real-world robotic manipulation tasks. H-GAR consistently outperforms state-of-the-art approaches, demonstrating particularly strong performance. Ablation studies further confirm the complementary roles and synergistic effects of the GOS and IAAR modules. Our main contributions are as follows:

\begin{itemize}
    \item We propose \textbf{H-GAR}, a goal-driven observation-action refinement framework for robotics that adopts a coarse-to-fine planning paradigm with integrated goal grounding and observation-action interaction.
    \item We propose \textbf{GOS} and \textbf{IAAR} to explicitly model the interaction between observation and action under the guidance of goal grounding: GOS synthesizes goal-aligned intermediate observations, and IAAR refines coarse actions using both historical actions and intermediate observations from GOS to produce coherent and task-consistent actions.
    \item We conduct comprehensive evaluations on both simulation and real-world robotic manipulation tasks, demonstrating the superior performance of H-GAR.
\end{itemize}

\section{2~~Related Work}

\paragraph{Goal-Conditioned Planning for Robotics.}
Recent works have shown that conditioning for goal states can significantly improve planning~\cite{luo2024grounding, bae2024tldr, gong2024goal, rens2025proposing, li2024optimus, li2025optimus, zhang2025falcon, lyu2025puma, shao2019multi, shen2024mome, chen2025SimpAgent}. LBP~\cite{liu2025efficient} learns the dynamics conditioned by latent goals and performs backward planning from the goal, improving the manipulation in the simulation. In the real world, SayCan~\cite{brohan2023can} combines language instructions with value grounding to guide robotic behavior, demonstrating strong performance in instruction-following tasks. Hu et al.\cite{hu2023planning} propose Planning Exploratory Goals (PEG) to encourage intrinsic goal-driven exploration, while Li et al.\cite{li2022hierarchical} decompose long-horizon tasks into goal-conditioned subtasks using hierarchical policies. 
Although these approaches leverage goal conditioning to improve efficiency or task decomposition, they often treat goal information as a constraint or policy input, without explicitly incorporating it into the perception-action generation process. 
In contrast, our method introduces a goal observation as a semantic anchor and explicitly integrates it into both observation synthesis and action refinement. 
\begin{figure*}[t]
    \centering
    
    \includegraphics[width=1\linewidth]{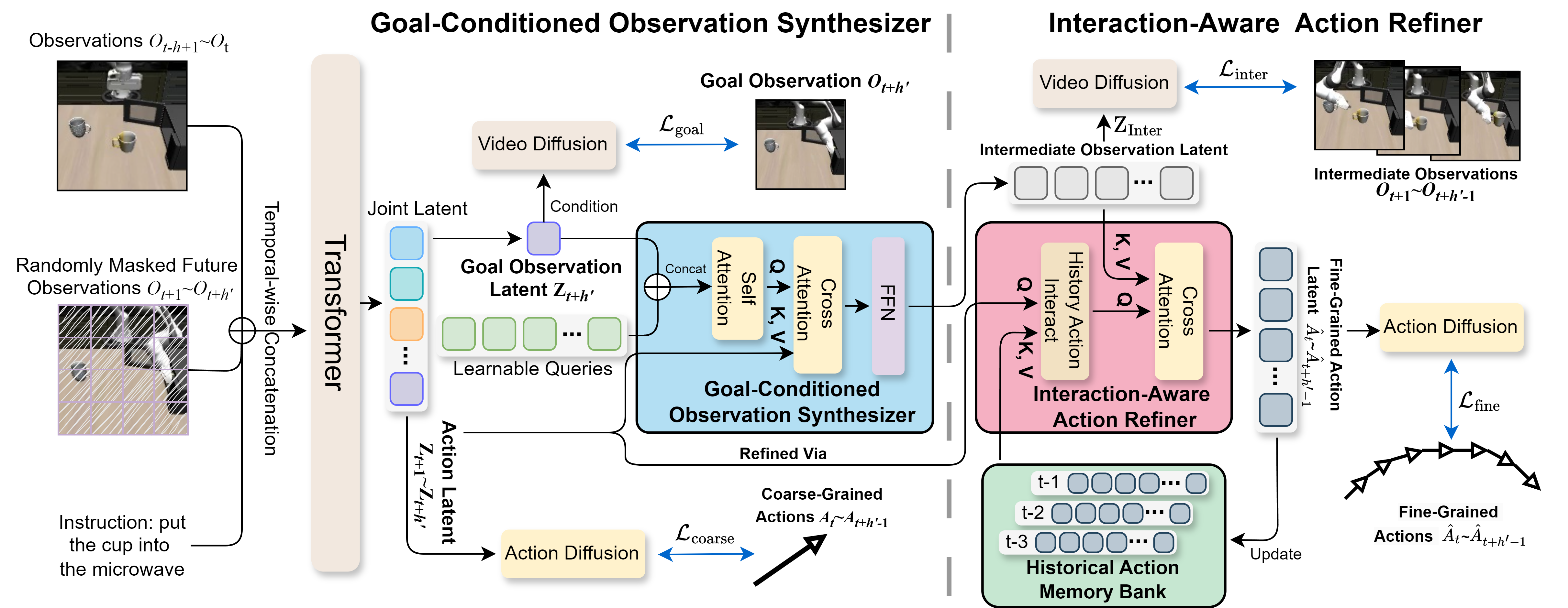}
    \vspace{-12pt}
\caption{\textbf{Overview of the H-GAR Framework.} H-GAR takes a task instruction and past observations $\{O_{t-h+1}, \dots, O_t\}$, along with masked future observations $\{O_{t+1}, \dots, O_{t+h'}\}$. The goal observation $O_{t+h'}$, which represents the final visual state to be achieved, is encoded and supervised to produce a semantic latent anchor that conditions the \textbf{Goal-Conditioned Observation Synthesizer (GOS)} to generate intermediate visual features. These features, together with a \textbf{Historical Action Memory Bank}, guide \textbf{Interaction-Aware Action Refiner (IAAR)} in denoising and refining latent actions. Both branches are trained via diffusion objectives to match ground-truth observations and actions. During training, future observations are randomly masked at the same positions across frames to avoid leakage, while inference starts from an empty image.}
\vspace{-10pt}
    \label{fig:method}
\end{figure*}
\paragraph{Future Observation Prediction for Robotic Manipulation.}
Anticipating future visual observations has proven to be a powerful mechanism for enhancing policy learning in robotics~\cite{hu2024video,lu2025h,huang2025enerverse,zhao2024vlmpc,xu2025vilp,zhu2025unified, shao2023detecting}. UWM~\cite{zhu2025unified} integrates action and video diffusion within a unified transformer, enhancing robustness and generalization in imitation learning. UVA~\cite{li2025unified} introduces the unified video action model, which jointly optimizes video and action predictions to achieve high accuracy. UP-VLA~\cite{zhang2025up} trains a unified vision-language-action model with joint multimodal understanding and future prediction objectives, enhancing both high-level semantic reasoning and low-level spatial understanding. PAD~\cite{guo2024prediction} unifies image prediction and robot action within a joint denoising process. However, these approaches often treat observation and action generation as parallel or loosely coupled processes, lacking structured interaction and goal conditioning. In contrast, our hierarchical coarse-to-fine framework couples observation and action bidirectionally, using goal-grounded prediction and intermediate visual states to enable temporally coherent, semantically aligned manipulation.
\section{3~~Methods}
\label{Method}
\subsection{3.1~~Preliminaries and Problem Formulation}
\label{Preliminaries}
In robotic manipulation, recent studies~\cite{cot, du2023learning, liang2024dreamitate, li2025unified} have demonstrated that predicting future visual observations can significantly enhance policy learning. A common paradigm in such frameworks is defined as follows:
Given a task instruction $I$ and a sequence of past visual observations $\{O_{t-h+1}, \dots, O_t\}$ over a history horizon $h$, the objective is to jointly predict a sequence of future action chunks $\{A_t, \dots, A_{t+h'-1}\}$ and the corresponding future observations $\{O_{t+1}, \dots, O_{t+h'}\}$ over a future horizon $h'$. Each action chunk $A_t \in \mathbb{R}^{K \times D}$ encodes $K$ consecutive low-level manipulation steps, where each step is represented as a $D$-dimensional action vector.

While effective, this paradigm presents two key limitations:
\textbf{(1) Goal-Agnostic Observation
Generation.}  
Without conditioning on a concrete goal observation, the model often generates future observations $\{O_{t+1}, \dots, O_{t+h'}\}$ that are visually plausible but semantically misaligned with the task objective implied by instruction $I$.
\textbf{(2) Implicit Observation-Action Modeling.}  
Actions $\{A_t, \dots, A_{t+h'-1}\}$ and observations $\{O_{t+1}, \dots, O_{t+h'}\}$ are typically predicted independently or with weak interaction, limiting the model’s ability to reason about their causal relationship and adapt to evolving visual feedback.

\subsection{3.2~~H-GAR: Overview}
To address these limitations, we propose \textbf{H-GAR}, which introduces explicit goal grounding and structured bidirectional interaction between observation and action. As illustrated in Fig.~\ref{fig:method}, instead of predicting the full observation sequence $\{O_{t+1}, \dots, O_{t+h'}\}$ at once, H-GAR decomposes it into a \textit{goal observation} $O_{t+h'}$—the final state after executing the action sequence—and \textit{intermediate observations} $\{O_{t+1}, \dots, O_{t+h'-1}\}$ capturing transitional visual feedback. To model their interaction with actions, we introduce the \textbf{Goal-Conditioned Observation Synthesizer (GOS)} and \textbf{Interaction-Aware Action Refiner (IAAR)}, which jointly enable coarse-to-fine action refinement.

Following prior works~\cite{chang2022maskgit, li2024autoregressive, li2025unified}, we encode each image observation into a sequence of $N$ latent tokens using a pretrained VAE~\cite{rombach2022high}, followed by a linear projection. 
Let $\mathbf{V}_{\mathcal{H}}$ and $\mathbf{V}_{\mathcal{F}}$ denote the tokenized representations from past and masked future observations, respectively. The task instruction $I$ is encoded into language embedding $T_I$ using a pretrained CLIP text encoder~\cite{radford2021learning}. All tokens are concatenated channel-wise to form a multimodal sequence, which is then processed by a Transformer encoder to produce joint latent representations for future steps:
\begin{equation}
\mathbf{Z}_{t+1:t+h'} = \text{Transformer}([\mathbf{V}_{\mathcal{H}},\ \mathbf{V}_{\mathcal{F}},\ T_I]).
\end{equation}

Here, $\mathbf{Z}_{t+1:t+h'} \in \mathbb{R}^{h' \times N \times D}$ represents the latent features for $h'$ future steps from $t{+}1$ to $t{+}h'$, where each step contains $N$ tokens with $D$-dimensional embeddings.

To generate a semantic goal anchor, we adopt a lightweight video diffusion decoder~\cite{li2024autoregressive} conditioned on the final latent $\mathbf{Z}_{t+h'}$. Each token $\mathbf{z}_i \in \mathbf{Z}_{t+h'} = \{\mathbf{z}_1, \dots, \mathbf{z}_N\}$ predicts a visual patch through a denoising process, which is decoded by a pretrained VAE to reconstruct the goal frame $O_{t+h'}$. The training objective minimizes the denoising error:
\begin{equation}
\mathcal{L}_{\text{goal}} = \mathbb{E}_{\epsilon, m} \left[ \frac{1}{N} \sum_{i=1}^N \left\| \epsilon_i - \epsilon_\phi(O^{(m)}_i \mid m, \mathbf{z}_i) \right\|_2^2 \right],
\label{eq:goal}
\end{equation}
where $O^{(m)}_i$ is the $i$-th noisy token of goal observation $O_{t+h'}$ at diffusion step $m$, and $\epsilon_\phi(\cdot)$ denotes the noise prediction network. 
Similarly, we aggregate the joint latents $\mathbf{Z}_{t+1:t+h'}$ to generate a coarse action sequence aligned with the goal observation. The training objective is defined as follows:
\begin{equation}
\mathcal{L}_{\text{coarse}} = \mathbb{E}_{\eta,\ m} \left[ \left\| 
\eta
- \eta_\theta({A}^{(m)} \mid m, \mathbf{Z}_{t+1:t+h'}) \right\|_2^2 \right],
\end{equation}
where ${A}^{(m)}$ is the ground-truth action chunk corrupted with noise at step $m$, and $\eta_\theta(\cdot)$ is the noise prediction network.

Next, given the above goal observation latent $\mathbf{Z}_{t+h'}$ and the coarse action latent $\mathbf{Z}_{t+1:t+h'}$, the \textbf{GOS} generates intermediate observation latent $\mathbf{Z}_{\text{Inter}}$ that bridges the high-level goal and low-level execution. Formally, we define it as:
\begin{equation}
 \mathbf{Z}_{\text{Inter}} = \textbf{GOS}(\mathbf{Z}_{t+h'},\mathbf{Z}_{t+1:t+h'}).
 \label{inter}
\end{equation}

We adopt a similar denoising objective as in goal prediction, using $\mathbf{Z}_{\text{Inter}}$ to reconstruct ground-truth intermediate observations corrupted with noise,  resulting in the loss \textbf{$\mathcal{L}_{\text{inter}}$}.
Finally, the \textbf{IAAR} updates the initial coarse action latent $\mathbf{Z}_{t+1:t+h'}$ by integrating intermediate observation features $\mathbf{Z}_{\text{Inter}}$ and the \textbf{Historical Action Memory Bank} $\mathcal{H}_t = [\hat{{A}_1}, \hat{{A}_2}, \dots, \hat{{A}}_{t-1}]$, producing a refined sequence $\hat{{A}}_{t:t+h'-1}$ that is temporally consistent and semantically aligned with the task goal:
\begin{equation}
\hat{{A}}_{t:t+h'-1} = \textbf{IAAR}(\mathbf{Z}_{t+1:t+h'},\ \mathbf{Z}_{\text{Inter}},\ \mathcal{H}_t).
\label{IAAR}
\end{equation}

Similarly, the refined action sequence $\hat{{A}}_{t+1:t+h'-1}$ is used to reconstruct the ground-truth actions through a diffusion-based denoising objective, yielding the loss \textbf{$\mathcal{L}_{\text{fine}}$}.
The overall training objective sums the losses introduced above: 
\begin{equation}
\mathcal{L}_{\text{total}} = \mathcal{L}_{\text{goal}} +  \mathcal{L}_{\text{coarse}} +  \mathcal{L}_{\text{inter}} +  \mathcal{L}_{\text{fine}}.
\end{equation}

\begin{table*}[t]

\caption{\textbf{Simulation Results.} 
Comparison of task success rates (SR) and ranks (RK) across both single-task and multi-task simulation settings. ``*'' denotes results reported in the original paper, while ``\dag'' denotes our reproduced results.}
\vspace{-5pt}
\label{tab:sim_results}
\centering
\footnotesize
\setlength{\tabcolsep}{15pt}
\begin{tabular}{c|cc|cc|cc}
\toprule
\textbf{Method}  
& \multicolumn{2}{c|}{\textbf{PushT}} & \multicolumn{2}{c|}{\textbf{PushT-M}} & \multicolumn{2}{c}{\textbf{Libero-10}} \\
\cmidrule(lr){2-3} \cmidrule(lr){4-5} \cmidrule(lr){6-7} 
& SR $\uparrow$ & RK $\downarrow$ & SR $\uparrow$ & RK $\downarrow$ & SR $\uparrow$ & RK $\downarrow$  \\
\midrule
Diffusion Policy-C*~\textit{[RSS'23]}~\cite{diff}        & 0.91 & 3  & 0.68 & 4  & 0.53 & 9 \\
Diffusion Policy-T*~\textit{[RSS'23]}~\cite{diff}        & 0.78 & 5  & 0.63 & 5  & 0.58 & 6 \\
UniPi*~\textit{[NeurIPS'23]}~\cite{du2023learning}   & 0.42 & 6 & 0.19 & 7   &- & - \\
OpenVLA*~\textit{[CoRL'24]}~\cite{openvla}             & 0.35 & 7  & 0.22& 6   & 0.54 & 8 \\
STAR*~\textit{[ICML'25]}~\cite{star}                   & -& -  & - & -  & 0.89 & 3 \\
CoT-VLA*~\textit{[CVPR'25]}~\cite{cotvla}              & - & - & - & -   & 0.69 & 5 \\
SpatialVLA*~\textit{[RSS'25]}~\cite{spatialvla}         & - &  -& - & -  & 0.56 &  7  \\
 PD-VLA\dag~\textit{[arXiv'25]}~\cite{pdvla}                & 0.82 &  4 & 0.71 &  3 &  0.92 &  2 \\
UVA\dag~\textit{[RSS'25]}~\cite{li2025unified} & 0.96 & 2  & 0.85& 2  & 0.89 & 3 \\

\hline
\textbf{H-GAR (Ours)}                                        & \textbf{0.99} & \textbf{1} & \textbf{0.90} & \textbf{1} &  \textbf{0.94} & \textbf{1} \\
\bottomrule
\end{tabular}
\vspace{-5pt}
\end{table*}

\begin{table*}[t]
\caption{\textbf{Real-World Experimental Results.} 
``*'' denotes results reported in the original paper, while ``\dag'' denotes our reproduced results. For long-horizon tasks (Object Placement, Drawer Manipulation), we report stage-wise completion rates.}
\vspace{-5pt}
\label{tab:alaho}
\centering
\footnotesize
\begin{tabular}{l|cc|ccc|c|c}
\toprule
\textbf{Method} & \multicolumn{2}{c|}{\textbf{Object Placement}} & \multicolumn{3}{c|}{\textbf{Drawer Manipulation}} & \textbf{Towel Folding} & \textbf{Mouse Arrangement}\\

& Cube$\rightarrow$Plate & +Toy$\rightarrow$Bowl & Open  & +Place & +Close &  &  \\
\midrule
VQ-BeT*~\cite{be} & 5/10 & 3/10 & 4/10 & 3/10 & 1/10 & - & - \\
QueST*~\cite{quest} & 6/10 & 4/10 & 3/10 & 1/10 & 0/10 & - & - \\
STAR*~\cite{star} & 8/10 & 6/10 & 6/10 & 4/10 & 3/10 & - & - \\
PD-VLA\dag~\cite{pdvla} & 8/10 & 7/10 & 6/10 & \textbf{6/10} & 4/10 & 6/10 & 4/10 \\
UVA\dag~\cite{li2025unified} & 7/10 & 6/10 & 6/10 & 5/10 & 3/10 & 5/10 & 3/10 \\
\textbf{H-GAR (Ours)} & \textbf{9/10} & \textbf{8/10} & \textbf{7/10} & \textbf{6/10} & \textbf{6/10} & \textbf{8/10} & \textbf{6/10} \\
\bottomrule
\end{tabular}
\vspace{-5pt}
\end{table*}

\begin{table}[t]
\centering
\footnotesize 
\setlength{\tabcolsep}{5pt}
\caption{\textbf{Observation Generation Results.}
Evaluation on simulated (Libero-10) and real-world (Mouse Arrangement) environments across different autoregressive steps. Lower FVD~\cite{unterthiner2018towards} reflects better visual fidelity.}
\vspace{-5pt}
\label{tab:libero10_fvd}
\begin{tabular}{l c c}
\toprule
\textbf{Method} & \textbf{Libero-10 $\downarrow$} & \textbf{Mouse Arrangement $\downarrow$} \\
\midrule
UniPi & 56.55  & 72.56\\
UVA (1 step) & 89.36     &59.32\\
UVA (8 steps) & 51.10      &32.78\\
\hline
\textbf{H-GAR (1 step)} & \textbf{86.76}  &\textbf{55.17}\\
\textbf{H-GAR (8 steps)} & \textbf{49.01} & \textbf{28.43}\\
\bottomrule
\end{tabular}
\vspace{-10pt}
\end{table}

\subsection{3.3~~H-GAR: Architectural Design of GOS and IAAR}
As outlined in Section~3.2, the hierarchical refinement process in H-GAR is implemented by the \textbf{GOS} and \textbf{IAAR} modules. This section details their architectural design and how they enable structured observation-action interaction.
\paragraph{Goal-Conditioned Observation Synthesis (GOS).}
Given the coarse action latent $\mathbf{Z}_{t+1:t+h'}$ and the goal observation latent $\mathbf{Z}_{t+h'}$, the GOS module synthesizes intermediate observation features that bridge high-level goal semantics and low-level action dynamics. We introduce a set of learnable queries $\mathbf{Q}_{\text{Inter}} \in \mathbb{R}^{(h'-1) \times N \times D}$, which are designed to represent latent intermediate frames.
To incorporate goal observation information, the queries are first updated via a self-attention module where the goal latent $\mathbf{Z}_{t+h'}$ is concatenated to the input:
\begin{equation}
\mathbf{Q}'_{\text{Inter}} = \textit{SelfAttn}([\mathbf{Q}_{\text{Inter}};\ \mathbf{Z}_{t+h'}]),
\end{equation}
where $\mathbf{Q}'_{\text{Inter}}$ denotes the updated queries obtained from the self-attention module $\textit{SelfAttn}(\cdot)$, and $[~;~]$ represents the concatenation operation.
Through this process, goal observation information is effectively aggregated into the queries. Subsequently, the updated queries $\mathbf{Q}'_{\text{Inter}}$ attend to the coarse action latent representations $\mathbf{Z}_{t+1:t+h'}$, which serve as keys and values, allowing the model to inject action-aware context into the intermediate observation synthesis. The resulting features are then refined through a feed-forward layer:
\begin{equation}
\mathbf{Z}_{\text{Inter}} = \text{FFN}\left(\textit{CrossAttn}(\mathbf{Q}'_{\text{Inter}},\ \mathbf{Z}_{t+1:t+h'})\right),
\end{equation}
where $\textit{CrossAttn}(\cdot)$ and $\text{FFN}(\cdot)$ denote the cross-attention and feed-forward layers, respectively. The output $\mathbf{Z}_{\text{Inter}}$ aligns with the definition in Eq.~\ref{inter}. This enables GOS to explicitly inject goal semantics and action context into intermediate observation synthesis, establishing a structured interaction between vision and action.
\paragraph{Interaction-Aware Action Refiner (IAAR).}
\begin{figure}[t]
    \centering
    \includegraphics[width=1.0\linewidth]{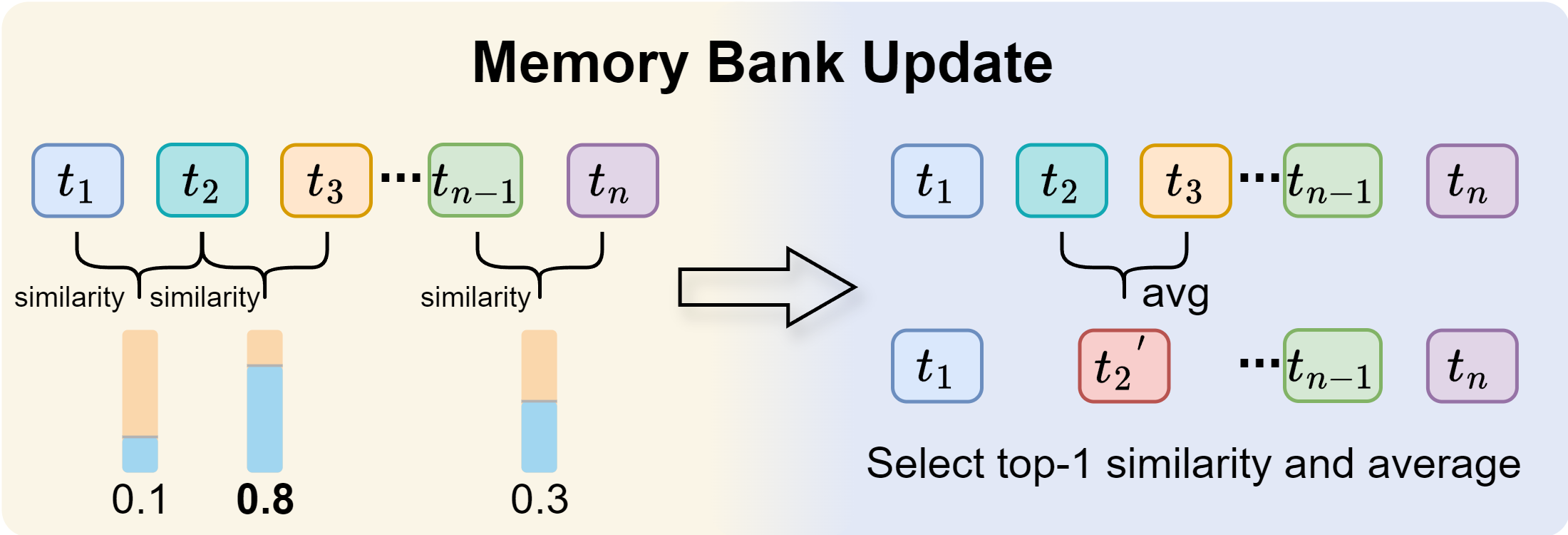}
    \caption{\textbf{Update Strategy of the Historical Action Memory Bank.} 
    When the memory exceeds a threshold, the most similar adjacent latents are merged via averaging.}
    \vspace{-10pt}
    \label{fig:memory}
\end{figure}
Building on the intermediate visual representations $\mathbf{Z}_{\text{Inter}}$ synthesized by GOS, we introduce the IAAR to refine coarse action latent $\mathbf{Z}_{t+1:t+h'}$. It integrates two sources of feedback: (1) temporal behavior priors from the \textbf{Historical Action Memory Bank} $\mathcal{H}_t = [\hat{{A}_1}, \hat{{A}_2}, \dots, \hat{{A}}_{t-1}]$, and (2) semantic visual feedback from the intermediate observation features $\mathbf{Z}_{\text{Inter}}$.

Firstly, we apply a history-interact layer $\textit{HisInter}(\cdot)$ to enhance temporal consistency and correct artifacts in the coarse action sequence. In this process, the \textbf{Historical Action Memory Bank} $\mathcal{H}_t$—which encodes history fine-grained action latent—serves as both \textit{key} and \textit{value}, while the coarse action latent ${Z}_{t+1:t+h'}$ acts as the \textit{query}.
\begin{equation}
\widetilde{{A}} = \textit{HisInter}(Q, K, V) = \text{Softmax}\left(\frac{Q K^\top}{\sqrt{d}}\right) V.
\end{equation}

Secondly, to inject observation-level context and improve semantic alignment with the synthesized intermediate observations, we refine $\widetilde{A}$ via a cross-attention layer $\textit{CrossAttn}(\cdot)$, where the intermediate observation latent $\mathbf{Z}_{\text{Inter}}$ serves as the \textit{key} and \textit{value} and $\widetilde{A}$ acts as the \textit{query}.
\begin{equation}
\hat{A}= \textit{CrossAttn}( \widetilde{A},\  \mathbf{Z}_{\text{Inter}}).
\end{equation}

The resulting $\hat{A}$ represents the refined fine-grained action latent $\hat{A}_{t:t+h'-1}$ (as defined in Eq.~\ref{IAAR}). The historical action memory bank then incorporates this latent and updates its contents following the strategy illustrated in Fig.~\ref{fig:memory}. Specifically, to maintain a compact yet informative memory $\mathcal{H}_t = [\hat{{A}_1}, \hat{{A}_2}, \dots, \hat{{A}}_{t+h'-1}]$, we apply a redundancy-aware compression when the memory exceeds a threshold. We first compute the cosine similarity between temporally adjacent action latent:
\begin{equation}
s_j = \cos(\hat{A_j},\ \hat{A}_{j+1}), \quad j \in [1, t+h'-2].
\end{equation}

Then we select the highest similarity pair across time and then average them to reduce memory length.

\section{4~~Experiments}
\subsection{4.1~~Experiment Settings}
\paragraph{Implementation details.} All experiments are conducted on 4 × A800 (80GB) GPUs. More implementation details are provided in \textbf{Appendix A}.

\paragraph{Simulation Benchmarks.}
Our simulation experiments include two settings: \textbf{Single-Task Evaluation} and \textbf{Multi-Task Evaluation}. In the single-task setting, we train and evaluate separate policies for each task, using \textbf{PushT}~\cite{chi2023diffusion, florence2022implicit} as a representative example. In the multi-task setting, a single policy is trained to handle multiple task goals. Following standard protocols, we evaluate on \textbf{PushT-M}~\cite{li2025unified} and \textbf{Libero-10}~\cite{liu2023libero}. Each task is tested in 50 diverse environments with different random seeds. Detailed descriptions of these simulation benchmarks are provided in \textbf{Appendix B.1}.
\paragraph{Real-World Setup.}
To validate our approach in the real world, we deploy it on the \textbf{Cobot Agilex ALOHA} platform across four manipulation tasks: \textbf{Object Placement, Drawer Manipulation, Towel Folding, and Mouse Arrangement}.  For the long-horizon tasks Object Placement and Drawer Manipulation, we additionally report stage-wise completion rates. Detailed descriptions are provided in \textbf{Appendix B.2}.

\paragraph{Baselines.} 
For robotic manipulation, we compare H-GAR with a range of state-of-the-art methods, such as Diffusion Policy-C/T, OpenVLA, SpatialVLA, and CoT-VLA. Diffusion Policy-C and T denote the CNN-based and Transformer-based variants of Diffusion Policy, respectively. For observation generation, we further compare H-GAR with UniPi and UVA, two state-of-the-art approaches in generative modeling for observation. Detailed descriptions of these baselines are provided in \textbf{Appendix B.3}.

\subsection{4.2~~Overall Performance}
\paragraph{Simulation Experimental Results.} 
As shown in Table~\ref{tab:sim_results}, H-GAR consistently achieves state-of-the-art performance across both single-task and multi-task scenarios, highlighting the robustness of our hierarchical, goal-driven framework in addressing both focused and diverse task distributions. Furthermore, Table~\ref{tab:libero10_fvd} shows that H-GAR achieves the lowest FVD scores under both 1-step and 8-step generation in both simulated and real-world settings, indicating superior visual fidelity. This demonstrates the advantage of our hierarchical generation strategy—first producing the goal observation and then synthesizing temporally coherent intermediate frames—over prior one-shot generation methods.
\paragraph{Real-World Experimental Results.}
We conduct real-world training and evaluation on four diverse manipulation tasks across robotic platform—Cobot Agilex ALOHA—to thoroughly assess the effectiveness and generalizability of our approach. As shown in Table~\ref{tab:alaho}, H-GAR achieves the highest stage-wise success rates across all tasks on the ALOHA platform, significantly outperforming prior methods in long-horizon settings such as Object Placement and Drawer Manipulation. Our method also excels in more dynamic and fine-grained tasks like Towel Folding and Mouse Arrangement, demonstrating robust visual grounding and precise control capabilities.

\begin{table}[t]
\caption{\textbf{Ablation Study on Model Components.} IAAR is evaluated with and without historical action memory bank.}
\vspace{-5pt}
\label{tab:model_components}
\centering
\footnotesize 
\setlength{\tabcolsep}{4pt}
\begin{tabular}{@{}c|cc|c c c@{}}
\toprule
\multicolumn{1}{c|}{\textbf{GOS}} & \multicolumn{2}{c|}{\textbf{IAAR}} & \multicolumn{1}{c}{\textbf{PushT}} & \multicolumn{1}{c}{\textbf{PushT-M}}  & \multicolumn{1}{c}{\textbf{Libero-10}} \\
 & w/o Bank & w/ Bank & &  & \\
\midrule
&     &     & 0.90	& 0.78   &0.85   \\
   \checkmark&     &     & 0.92	& 0.82   &0.89   \\
\checkmark& \checkmark &  &0.96  &0.87   &0.91  \\
\checkmark&    & \checkmark &\textbf{0.99}  &\textbf{0.90}    &\textbf{0.94}  \\
\midrule
\end{tabular}
\vspace{-5pt}
\end{table}

\begin{table}[t]
\centering
\footnotesize 
\setlength{\tabcolsep}{6pt}
\caption{\textbf{Ablation Study on Goal-Conditioned Observation Strategy.} 
We compare input strategies for GOS: using a single random frame,  multiple uniformly sampled frames, and the goal frame.}
\vspace{-5pt}
\label{tab:ablation_observation}
\begin{tabular}{lccc}
\toprule
\textbf{Selection Strategy} & \textbf{PushT} & \textbf{PushT-M} &  \textbf{Libero-10} \\
\midrule
Uniform Multi-Frame     & 0.96 & 0.83  & 0.91 \\
Single Random Frame     & 0.95 & 0.86  & 0.88 \\
Goal Frame Only       & \textbf{0.99} & \textbf{0.90}  & \textbf{0.94} \\
\bottomrule
\end{tabular}
\vspace{-10pt}
\end{table}
\begin{table}[t]
\centering
\footnotesize
\setlength{\tabcolsep}{12pt}
\caption{\textbf{Ablation Study on Historical Action Memory Bank.} 
We evaluate the impact of memory bank size and update strategies. 
Compared strategies include Random, FIFO (First-In, First-Out), and Similarity (Ours), which discards the most similar entry to promote diversity.}
\vspace{-5pt}
\label{tab:ablation_actionbank}
\begin{tabular}{lccc}
\toprule
\textbf{Settings} & \textbf{PushT} & \textbf{PushT-M}  & \textbf{Libero-10} \\
\midrule
\multicolumn{4}{c}{\textbf{\textit{Action Memory Bank Size}}} \\
\midrule
8   & 0.96 & 0.88 & 0.90 \\
16  & 0.97 & 0.88  & 0.91 \\
32  & 0.98 & \textbf{0.90}  & \textbf{0.94} \\
64  & \textbf{0.99} & 0.89  & 0.92 \\
\midrule
\multicolumn{4}{c}{\textbf{\textit{Update Strategy}}} \\
\midrule
Random     & 0.95 & 0.86  & 0.89 \\
FIFO  & 0.97 & 0.88  & 0.91 \\
Similarity & \textbf{0.99} & \textbf{0.90} & \textbf{0.94} \\
\bottomrule
\end{tabular}
\vspace{-5pt}
\end{table}
\begin{figure}[t]
    \centering
    \includegraphics[width=1\linewidth]{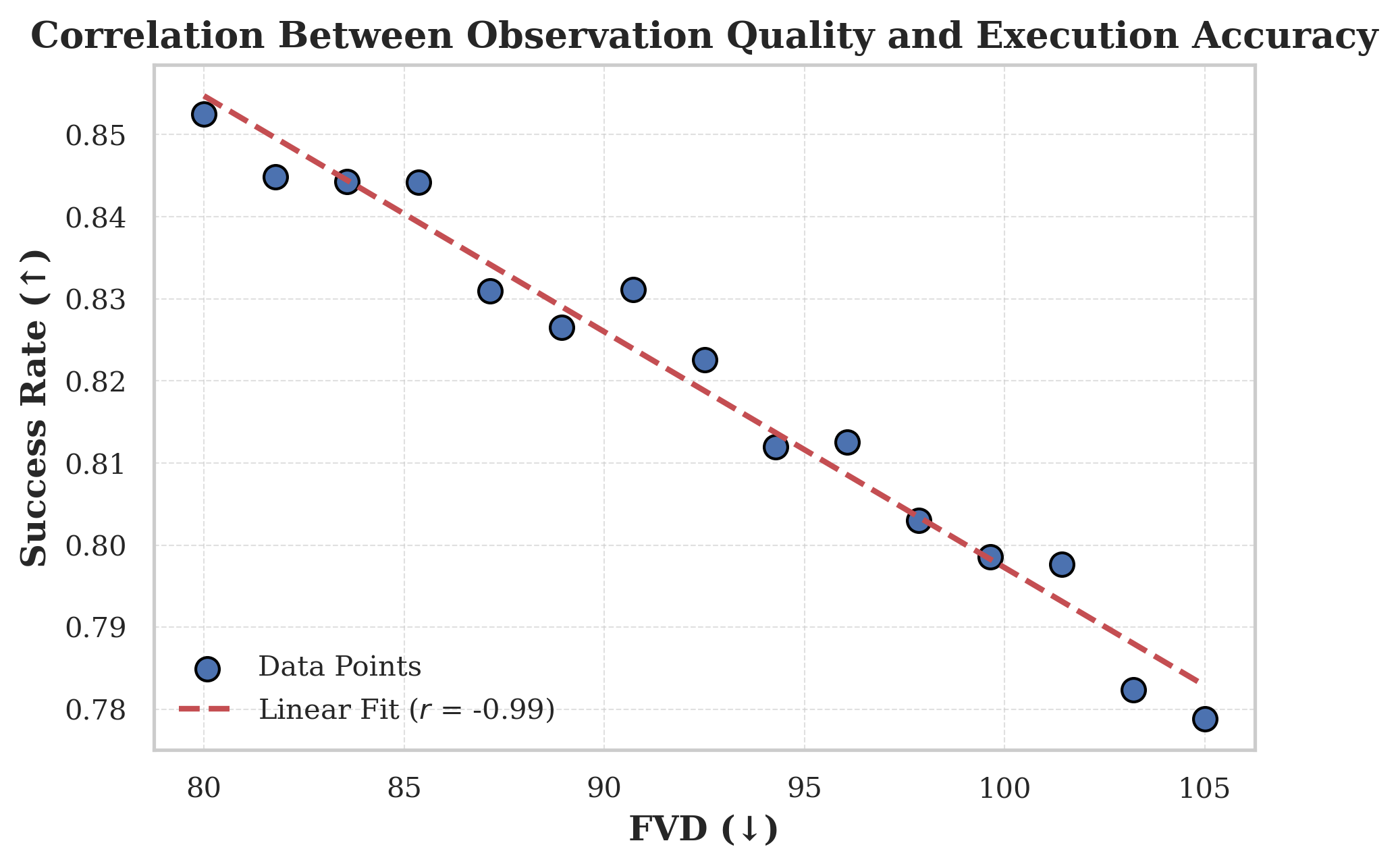}
    \vspace{-18pt}
 \caption{\textbf{FVD vs. Success Rate on Libero-10.} 
A strong negative correlation is observed, indicating that better observation generation quality (lower FVD) leads to higher task success rates.}
    \label{fig:fvd}
    \vspace{-12pt}
\end{figure}

\begin{figure*}[t]
\centering
\includegraphics[width=1.0\textwidth]{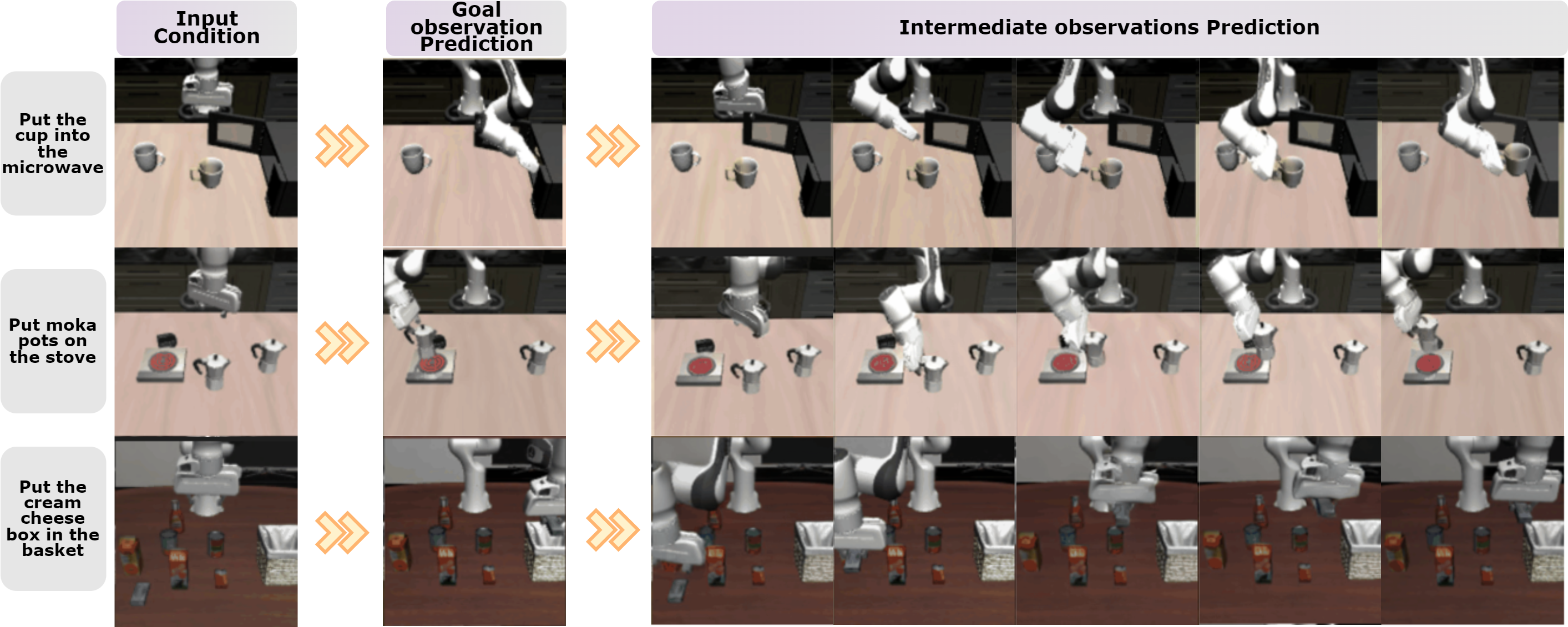}
\vspace{-15pt}
\caption{\textbf{Visualization of Goal and Intermediate Observation Prediction.} Given a task instruction and initial scene, H-GAR first predicts the goal observation, followed by intermediate frames capturing temporally coherent transitions.}
\vspace{-5pt}
\label{vis_gene}
\end{figure*}
\begin{figure*}[t]
\centering
\includegraphics[width=1.0\textwidth]{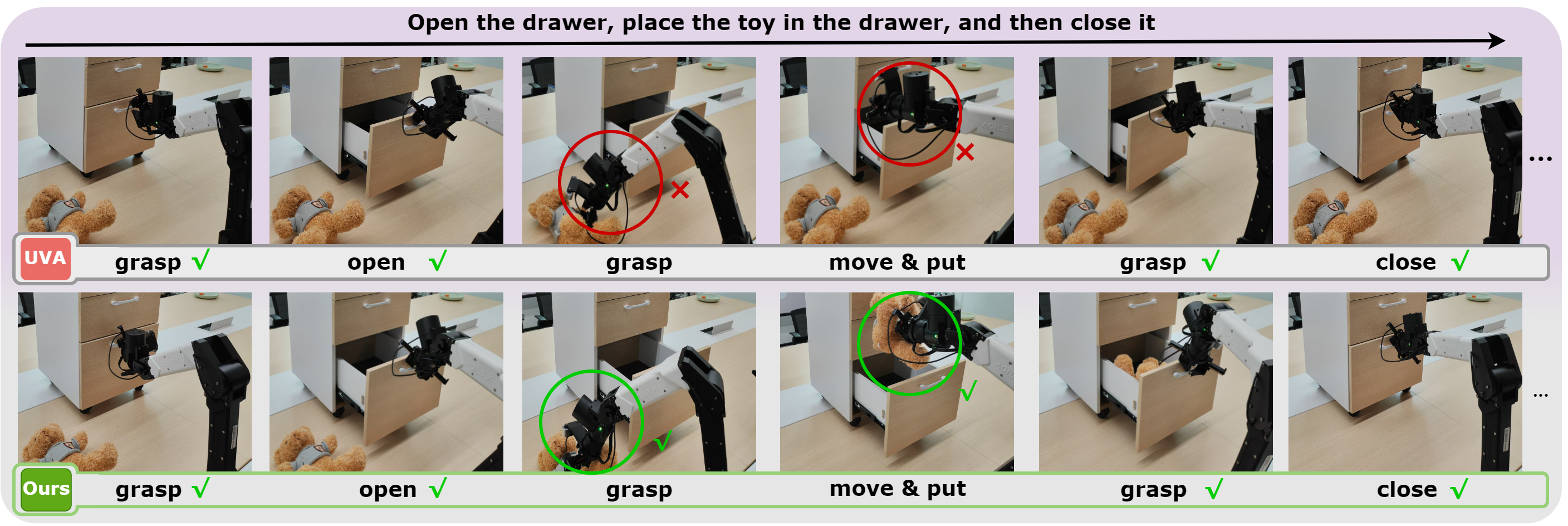}
\vspace{-15pt}
\caption{\textbf{Visualization of Real-World Execution Comparison.} 
Given the long-horizon instruction,
H-GAR successfully completes all stages of the manipulation task, achieving coherent transitions, 
while UVA fails at multiple critical steps, such as grasping and object placement.}
\label{vis_robo}
\vspace{-12pt}
\end{figure*}
\subsection{4.3~~Ablation Studies}
\paragraph{Ablation on Model Components.} 
As shown in Table~\ref{tab:model_components}, both the Goal-Conditioned Observation Selection (GOS) and the Interaction-Aware Action Refinement (IAAR) modules contribute significantly to overall performance. In particular, the historical action memory in IAAR consistently improves performance, highlighting the benefit of using temporal action history for policy refinement.
\paragraph{Ablation on Goal-Guided Observation Generation strategy.} 
Table~\ref{tab:ablation_observation} presents a comparison of different input strategies for the GOS module. 
Across all tasks, conditioning on the predicted goal observation consistently yields the highest success rates, 
surpassing both random single-frame and uniform multi-frame baselines. 
This highlights the critical role of explicitly generating a goal observation, 
which serves as a strong semantic anchor to align synthesized intermediate observations with the final task objective. 
In contrast, while other strategies provide additional local visual context, 
they lack explicit task-driven guidance, often leading to semantically plausible but goal-irrelevant rollouts. 
By grounding the generation process in a clear, high-level visual target, 
our \emph{goal-first} design ensures temporally coherent transitions aligned with the intended outcome, 
highlighting the necessity of this architectural choice. 
\paragraph{Ablation on Historical Action Memory
Bank.}
Table~\ref{tab:ablation_actionbank} explores the design choices of the memory bank in IAAR. Increasing the memory size generally improves performance. However, overly long memory may introduce distracting or outdated historical actions, which can negatively impact policy execution. In terms of update strategies, our proposed similarity-based eviction method outperforms both FIFO and random baselines. The FIFO strategy (first-in, first-out) employs a queue-based mechanism that removes the oldest entry upon insertion. In contrast, the similarity-based strategy enhances diversity by removing the most similar entry, preserving more distinct historical contexts.
\paragraph{The Correlation between Observation Generation and Task Success.}
Figure~\ref{fig:fvd} shows a strong negative correlation between Fréchet Video Distance (FVD) and task success rate on Libero-10, indicating that higher-quality observation generation leads to better downstream execution. These results validate the effectiveness of our hierarchical, goal-driven design and emphasize the importance of aligning observation generation with action-level reasoning.

\subsection{4.4~~Qualitative Analysis}
\paragraph{Visualization of Observation Prediction.}
Figure~\ref{vis_gene} illustrates the hierarchical prediction process of H-GAR. Given a task instruction and an initial observation, the model first generates a goal observation that anchors the high-level intent, followed by a sequence of intermediate frames that depict temporally coherent transitions toward task completion. This coarse-to-fine structure enables long-horizon planning by explicitly grounding sub-goals.

\paragraph{Visualization of Real-World Comparison.}
Figure~\ref{vis_robo} shows a real-world comparison between H-GAR and UVA on a complex multi-stage manipulation task. While UVA fails at key stages such as object placement and grasping, H-GAR successfully completes all subtasks—including drawer opening, grasping, toy placement, and drawer closing—with high accuracy and temporal consistency.  These visualizations underscore H-GAR’s ability to maintain task progression and spatial reasoning under real-world uncertainties, further validating the effectiveness of its hierarchical, goal-driven design. Additional visualizations are provided in \textbf{Appendix C}, and \textbf{real-world execution videos are available in the supplementary material.}

\section{5~~Conclusion}
In this work, we presented H-GAR, a hierarchical goal-driven framework for robotic manipulation that integrates goal-conditioned observation generation with interaction-aware action refinement. By first predicting the goal observation and subsequently generating temporally coherent intermediate transitions, H-GAR facilitates structured long-horizon planning and execution. Extensive experiments across both simulation and real-world environments demonstrate that our method achieves state-of-the-art performance. Comprehensive ablation studies further highlight the critical role of both the Goal Observation Synthesizer (GOS) in providing goal-aligned visual guidance and the Interaction-Aware Action Refiner (IAAR) in facilitating precise and consistent action generation.

\bibliography{aaai2026}

\begin{thebibliography}{61}
\providecommand{\natexlab}[1]{#1}

\bibitem[{Bae, Park, and Lee(2024)}]{bae2024tldr}
Bae, J.; Park, K.; and Lee, Y. 2024.
\newblock Tldr: Unsupervised goal-conditioned rl via temporal distance-aware representations.
\newblock \emph{arXiv preprint arXiv:2407.08464}.

\bibitem[{Bharadhwaj et~al.(2024)Bharadhwaj, Vakil, Sharma, Gupta, Tulsiani, and Kumar}]{roboagent}
Bharadhwaj, H.; Vakil, J.; Sharma, M.; Gupta, A.; Tulsiani, S.; and Kumar, V. 2024.
\newblock Roboagent: Generalization and efficiency in robot manipulation via semantic augmentations and action chunking.
\newblock In \emph{2024 IEEE International Conference on Robotics and Automation (ICRA)}, 4788--4795. IEEE.

\bibitem[{Brohan et~al.(2023)Brohan, Chebotar, Finn, Hausman, Herzog, Ho, Ibarz, Irpan, Jang, Julian et~al.}]{brohan2023can}
Brohan, A.; Chebotar, Y.; Finn, C.; Hausman, K.; Herzog, A.; Ho, D.; Ibarz, J.; Irpan, A.; Jang, E.; Julian, R.; et~al. 2023.
\newblock Do as i can, not as i say: Grounding language in robotic affordances.
\newblock In \emph{Conference on robot learning}, 287--318. PMLR.

\bibitem[{Chang et~al.(2022)Chang, Zhang, Jiang, Liu, and Freeman}]{chang2022maskgit}
Chang, H.; Zhang, H.; Jiang, L.; Liu, C.; and Freeman, W.~T. 2022.
\newblock Maskgit: Masked generative image transformer.
\newblock In \emph{Proceedings of the IEEE/CVF conference on computer vision and pattern recognition}, 11315--11325.

\bibitem[{Chen et~al.(2025)Chen, Zhou, Shao, Lyu, Zhou, Wang, Li, Li, Qi, and Nie}]{chen2025SimpAgent}
Chen, G.; Zhou, X.; Shao, R.; Lyu, Y.; Zhou, K.; Wang, S.; Li, W.; Li, Y.; Qi, Z.; and Nie, L. 2025.
\newblock Less is More: Empowering GUI Agent with Context-Aware Simplification.
\newblock In \emph{Proceedings of the IEEE/CVF International Conference on Computer Vision}.

\bibitem[{Chi et~al.(2023{\natexlab{a}})Chi, Luo, Duan, Handa, and Fox}]{Chi2023diffusion_policy}
Chi, C.; Luo, Y.; Duan, Y.; Handa, A.; and Fox, D. 2023{\natexlab{a}}.
\newblock Diffusion Policy: Visuomotor Policy Learning via Action Diffusion.
\newblock \emph{The International Journal of Robotics Research}.

\bibitem[{Chi et~al.(2023{\natexlab{b}})Chi, Xu, Feng, Cousineau, Du, Burchfiel, Tedrake, and Song}]{diff}
Chi, C.; Xu, Z.; Feng, S.; Cousineau, E.; Du, Y.; Burchfiel, B.; Tedrake, R.; and Song, S. 2023{\natexlab{b}}.
\newblock Diffusion policy: Visuomotor policy learning via action diffusion.
\newblock \emph{The International Journal of Robotics Research}, 02783649241273668.

\bibitem[{Chi et~al.(2023{\natexlab{c}})Chi, Xu, Feng, Cousineau, Du, Burchfiel, Tedrake, and Song}]{chi2023diffusion}
Chi, C.; Xu, Z.; Feng, S.; Cousineau, E.; Du, Y.; Burchfiel, B.; Tedrake, R.; and Song, S. 2023{\natexlab{c}}.
\newblock Diffusion policy: Visuomotor policy learning via action diffusion.
\newblock \emph{The International Journal of Robotics Research}, 02783649241273668.

\bibitem[{Du et~al.(2023{\natexlab{a}})Du, Dai, Zeng, Handa, Fox, Tenenbaum, Freeman, and Wu}]{unipi}
Du, Y.; Dai, X.; Zeng, A.; Handa, A.; Fox, D.; Tenenbaum, J.~B.; Freeman, W.~T.; and Wu, J. 2023{\natexlab{a}}.
\newblock Learning Universal Policies via Text-Guided Video Generation.
\newblock In \emph{Advances in Neural Information Processing Systems (NeurIPS)}, volume~36, 9156--9172.

\bibitem[{Du et~al.(2023{\natexlab{b}})Du, Yang, Dai, Dai, Nachum, Tenenbaum, Schuurmans, and Abbeel}]{du2023learning}
Du, Y.; Yang, S.; Dai, B.; Dai, H.; Nachum, O.; Tenenbaum, J.; Schuurmans, D.; and Abbeel, P. 2023{\natexlab{b}}.
\newblock Learning universal policies via text-guided video generation.
\newblock \emph{Advances in neural information processing systems}, 36: 9156--9172.

\bibitem[{Esser, Rombach, and Ommer(2021)}]{esser2021VQGAN}
Esser, P.; Rombach, R.; and Ommer, B. 2021.
\newblock Taming Transformers for High-Resolution Image Synthesis.
\newblock In \emph{Proceedings of the IEEE/CVF Conference on Computer Vision and Pattern Recognition (CVPR)}, 12873--12883.

\bibitem[{Florence et~al.(2022)Florence, Lynch, Zeng, Ramirez, Wahid, Downs, Wong, Lee, Mordatch, and Tompson}]{florence2022implicit}
Florence, P.; Lynch, C.; Zeng, A.; Ramirez, O.~A.; Wahid, A.; Downs, L.; Wong, A.; Lee, J.; Mordatch, I.; and Tompson, J. 2022.
\newblock Implicit behavioral cloning.
\newblock In \emph{Conference on robot learning}, 158--168. PMLR.

\bibitem[{Gong et~al.(2024)Gong, Dawei, Xu, Ding, and Wang}]{gong2024goal}
Gong, X.; Dawei, F.; Xu, K.; Ding, B.; and Wang, H. 2024.
\newblock Goal-conditioned on-policy reinforcement learning.
\newblock \emph{Advances in neural information processing systems}, 37: 45975--46001.

\bibitem[{Guo et~al.(2024)Guo, Hu, Zhang, Wang, Chen, Lu, and Chen}]{guo2024prediction}
Guo, Y.; Hu, Y.; Zhang, J.; Wang, Y.-J.; Chen, X.; Lu, C.; and Chen, J. 2024.
\newblock Prediction with action: Visual policy learning via joint denoising process.
\newblock In \emph{The Thirty-eighth Annual Conference on Neural Information Processing Systems}.

\bibitem[{Hao et~al.(2025)Hao, Qi, Rui, Xiang, Yinchuan, Jianye, and Liqiang}]{star}
Hao, L.; Qi, L.; Rui, S.; Xiang, D.; Yinchuan, L.; Jianye, H.; and Liqiang, N. 2025.
\newblock STAR: Learning Diverse Robot Skill Abstractions through Rotation-Augmented Vector Quantization.
\newblock \emph{International Conference on Machine Learning (ICML)}.

\bibitem[{Hu et~al.(2023)Hu, Chang, Rybkin, and Jayaraman}]{hu2023planning}
Hu, E.~S.; Chang, R.; Rybkin, O.; and Jayaraman, D. 2023.
\newblock Planning goals for exploration.
\newblock \emph{arXiv preprint arXiv:2303.13002}.

\bibitem[{Hu et~al.(2024)Hu, Guo, Wang, Chen, Wang, Zhang, Sreenath, Lu, and Chen}]{hu2024video}
Hu, Y.; Guo, Y.; Wang, P.; Chen, X.; Wang, Y.-J.; Zhang, J.; Sreenath, K.; Lu, C.; and Chen, J. 2024.
\newblock Video prediction policy: A generalist robot policy with predictive visual representations.
\newblock \emph{arXiv preprint arXiv:2412.14803}.

\bibitem[{Huang et~al.(2025)Huang, Chen, Zhou, Chen, Jiang, Hu, Liao, Gao, Li, Yao et~al.}]{huang2025enerverse}
Huang, S.; Chen, L.; Zhou, P.; Chen, S.; Jiang, Z.; Hu, Y.; Liao, Y.; Gao, P.; Li, H.; Yao, M.; et~al. 2025.
\newblock Enerverse: Envisioning embodied future space for robotics manipulation.
\newblock \emph{arXiv preprint arXiv:2501.01895}.

\bibitem[{Kim et~al.(2024{\natexlab{a}})Kim, Luo, Xu, Duan, Luo, Handa, and Fox}]{Kim2024openvla}
Kim, M.~J.; Luo, Y.; Xu, W.; Duan, Y.; Luo, Y.; Handa, A.; and Fox, D. 2024{\natexlab{a}}.
\newblock OpenVLA: An Open-Source Vision-Language-Action Model.
\newblock \emph{arXiv preprint arXiv:2406.09246}.

\bibitem[{Kim et~al.(2024{\natexlab{b}})Kim, Pertsch, Karamcheti, Xiao, Balakrishna, Nair, Rafailov, Foster, Lam, Sanketi et~al.}]{openvla}
Kim, M.~J.; Pertsch, K.; Karamcheti, S.; Xiao, T.; Balakrishna, A.; Nair, S.; Rafailov, R.; Foster, E.; Lam, G.; Sanketi, P.; et~al. 2024{\natexlab{b}}.
\newblock Openvla: An open-source vision-language-action model.
\newblock \emph{arXiv preprint arXiv:2406.09246}.

\bibitem[{Lee et~al.(2024{\natexlab{a}})Lee, Wang, Etukuru, Kim, Shafiullah, and Pinto}]{be}
Lee, S.; Wang, Y.; Etukuru, H.; Kim, H.~J.; Shafiullah, N. M.~M.; and Pinto, L. 2024{\natexlab{a}}.
\newblock Behavior generation with latent actions.
\newblock \emph{arXiv preprint arXiv:2403.03181}.

\bibitem[{Lee et~al.(2024{\natexlab{b}})Lee, Wang, Etukuru, Kim, Shafiullah, and Pinto}]{vqbet}
Lee, S.; Wang, Y.; Etukuru, H.; Kim, H.~J.; Shafiullah, N. M.~M.; and Pinto, L. 2024{\natexlab{b}}.
\newblock Behavior generation with latent actions.
\newblock \emph{arXiv preprint arXiv:2403.03181}.

\bibitem[{Li et~al.(2022)Li, Tang, Tomizuka, and Zhan}]{li2022hierarchical}
Li, J.; Tang, C.; Tomizuka, M.; and Zhan, W. 2022.
\newblock Hierarchical planning through goal-conditioned offline reinforcement learning.
\newblock \emph{IEEE Robotics and Automation Letters}, 7(4): 10216--10223.

\bibitem[{Li et~al.(2024{\natexlab{a}})Li, Liang, Wang, Luo, Chen, Liao, Wei, Deng, Xu, Zhang et~al.}]{cogact}
Li, Q.; Liang, Y.; Wang, Z.; Luo, L.; Chen, X.; Liao, M.; Wei, F.; Deng, Y.; Xu, S.; Zhang, Y.; et~al. 2024{\natexlab{a}}.
\newblock Cogact: A foundational vision-language-action model for synergizing cognition and action in robotic manipulation.
\newblock \emph{arXiv preprint arXiv:2411.19650}.

\bibitem[{Li et~al.(2025{\natexlab{a}})Li, Gao, Sadigh, and Song}]{li2025unified}
Li, S.; Gao, Y.; Sadigh, D.; and Song, S. 2025{\natexlab{a}}.
\newblock Unified video action model.
\newblock \emph{arXiv preprint arXiv:2503.00200}.

\bibitem[{Li et~al.(2024{\natexlab{b}})Li, Wang, Zhai, Hou, He, Song, Ren, Belongie, Lucic, and Ma}]{li2024autoregressive}
Li, T.; Wang, Z.; Zhai, X.; Hou, L.; He, Y.; Song, Y.; Ren, M.; Belongie, S.; Lucic, M.; and Ma, X. 2024{\natexlab{b}}.
\newblock Autoregressive Image Generation without Vector Quantization.
\newblock In \emph{Advances in Neural Information Processing Systems (NeurIPS)}, volume~37, 56424--56445.

\bibitem[{Li et~al.(2025{\natexlab{b}})Li, Hu, Shao, Shen, and Nie}]{li2025lion}
Li, W.; Hu, B.; Shao, R.; Shen, L.; and Nie, L. 2025{\natexlab{b}}.
\newblock Lion-fs: Fast \& slow video-language thinker as online video assistant.
\newblock In \emph{Proceedings of the Computer Vision and Pattern Recognition Conference}, 3240--3251.

\bibitem[{Li et~al.(2025{\natexlab{c}})Li, Zhang, Shao, He, and Nie}]{li2025cogvla}
Li, W.; Zhang, R.; Shao, R.; He, J.; and Nie, L. 2025{\natexlab{c}}.
\newblock Cogvla: Cognition-aligned vision-language-action model via instruction-driven routing \& sparsification.
\newblock \emph{arXiv preprint arXiv:2508.21046}.

\bibitem[{Li et~al.(2024{\natexlab{c}})Li, Xie, Shao, Chen, Jiang, and Nie}]{li2024optimus}
Li, Z.; Xie, Y.; Shao, R.; Chen, G.; Jiang, D.; and Nie, L. 2024{\natexlab{c}}.
\newblock Optimus-1: Hybrid Multimodal Memory Empowered Agents Excel in Long-Horizon Tasks.
\newblock In \emph{Advances in Neural Information Processing Systems}, volume~37, 49881--49913.

\bibitem[{Li et~al.(2025{\natexlab{d}})Li, Xie, Shao, Chen, Jiang, and Nie}]{li2025optimus}
Li, Z.; Xie, Y.; Shao, R.; Chen, G.; Jiang, D.; and Nie, L. 2025{\natexlab{d}}.
\newblock Optimus-2: Multimodal minecraft agent with goal-observation-action conditioned policy.
\newblock In \emph{Proceedings of the Computer Vision and Pattern Recognition Conference}, 9039--9049.

\bibitem[{Liang et~al.(2024)Liang, Liu, Ozguroglu, Sudhakar, Dave, Tokmakov, Song, and Vondrick}]{liang2024dreamitate}
Liang, J.; Liu, R.; Ozguroglu, E.; Sudhakar, S.; Dave, A.; Tokmakov, P.; Song, S.; and Vondrick, C. 2024.
\newblock Dreamitate: Real-world visuomotor policy learning via video generation.
\newblock \emph{arXiv preprint arXiv:2406.16862}.

\bibitem[{Liu et~al.(2023)Liu, Zhu, Gao, Feng, Liu, Zhu, and Stone}]{liu2023libero}
Liu, B.; Zhu, Y.; Gao, C.; Feng, Y.; Liu, Q.; Zhu, Y.; and Stone, P. 2023.
\newblock Libero: Benchmarking knowledge transfer for lifelong robot learning.
\newblock \emph{Advances in Neural Information Processing Systems}, 36: 44776--44791.

\bibitem[{Liu et~al.(2025)Liu, Niu, Wang, Zheng, Zheng, Ou, Hu, Li, and Zhan}]{liu2025efficient}
Liu, D.; Niu, H.; Wang, Z.; Zheng, J.; Zheng, Y.; Ou, Z.; Hu, J.; Li, J.; and Zhan, X. 2025.
\newblock Efficient Robotic Policy Learning via Latent Space Backward Planning.
\newblock \emph{arXiv preprint arXiv:2505.06861}.

\bibitem[{Lu et~al.(2025)Lu, Tian, Yuan, Wang, Hua, Xue, and Xu}]{lu2025h}
Lu, Y.; Tian, Y.; Yuan, Z.; Wang, X.; Hua, P.; Xue, Z.; and Xu, H. 2025.
\newblock {H\textsuperscript{3}DP}: Triply-Hierarchical Diffusion Policy for Visuomotor Learning.
\newblock \emph{arXiv preprint arXiv:2505.07819}.

\bibitem[{Luo and Du(2024)}]{luo2024grounding}
Luo, Y.; and Du, Y. 2024.
\newblock Grounding video models to actions through goal conditioned exploration.
\newblock \emph{arXiv preprint arXiv:2411.07223}.

\bibitem[{Lyu et~al.(2025)Lyu, Shao, Chen, Zhu, Guan, and Nie}]{lyu2025puma}
Lyu, Y.; Shao, R.; Chen, G.; Zhu, Y.; Guan, W.; and Nie, L. 2025.
\newblock PUMA: Layer-Pruned Language Model for Efficient Unified Multimodal Retrieval with Modality-Adaptive Learning.
\newblock In \emph{Proceedings of the 33nd ACM International Conference on Multimedia}.

\bibitem[{Mete et~al.(2024)Mete, Xue, Wilcox, Chen, and Garg}]{quest}
Mete, A.; Xue, H.; Wilcox, A.; Chen, Y.; and Garg, A. 2024.
\newblock QueST: Self-Supervised Skill Abstractions for Learning Continuous Control.
\newblock arXiv:2407.15840.

\bibitem[{Qu et~al.(2025)Qu, Song, Chen, Yao, Ye, Ding, Wang, Gu, Zhao, Wang et~al.}]{spatialvla}
Qu, D.; Song, H.; Chen, Q.; Yao, Y.; Ye, X.; Ding, Y.; Wang, Z.; Gu, J.; Zhao, B.; Wang, D.; et~al. 2025.
\newblock SpatialVLA: Exploring Spatial Representations for Visual-Language-Action Model.
\newblock \emph{arXiv preprint arXiv:2501.15830}.

\bibitem[{Radford et~al.(2021)Radford, Kim, Hallacy, Ramesh, Goh, Agarwal, Sastry, Askell, Mishkin, Clark et~al.}]{radford2021learning}
Radford, A.; Kim, J.~W.; Hallacy, C.; Ramesh, A.; Goh, G.; Agarwal, S.; Sastry, G.; Askell, A.; Mishkin, P.; Clark, J.; et~al. 2021.
\newblock Learning transferable visual models from natural language supervision.
\newblock In \emph{International conference on machine learning}, 8748--8763. PmLR.

\bibitem[{Rens(2025)}]{rens2025proposing}
Rens, G.~B. 2025.
\newblock Proposing Hierarchical Goal-Conditioned Policy Planning in Multi-Goal Reinforcement Learning.
\newblock \emph{arXiv preprint arXiv:2501.01727}.

\bibitem[{Rombach et~al.(2022)Rombach, Blattmann, Lorenz, Esser, and Ommer}]{rombach2022high}
Rombach, R.; Blattmann, A.; Lorenz, D.; Esser, P.; and Ommer, B. 2022.
\newblock High-resolution image synthesis with latent diffusion models.
\newblock In \emph{Proceedings of the IEEE/CVF conference on computer vision and pattern recognition}, 10684--10695.

\bibitem[{Shao et~al.(2019)Shao, Lan, Li, and Yuen}]{shao2019multi}
Shao, R.; Lan, X.; Li, J.; and Yuen, P.~C. 2019.
\newblock Multi-adversarial discriminative deep domain generalization for face presentation attack detection.
\newblock In \emph{Proceedings of the IEEE/CVF conference on computer vision and pattern recognition}, 10023--10031.

\bibitem[{Shao, Wu, and Liu(2023)}]{shao2023detecting}
Shao, R.; Wu, T.; and Liu, Z. 2023.
\newblock Detecting and grounding multi-modal media manipulation.
\newblock In \emph{Proceedings of the IEEE/CVF Conference on Computer Vision and Pattern Recognition}, 6904--6913.

\bibitem[{Shao et~al.(2024)Shao, Wu, Wu, Nie, and Liu}]{shao2024detecting}
Shao, R.; Wu, T.; Wu, J.; Nie, L.; and Liu, Z. 2024.
\newblock Detecting and grounding multi-modal media manipulation and beyond.
\newblock \emph{IEEE Transactions on Pattern Analysis and Machine Intelligence}.

\bibitem[{Shen et~al.(2024)Shen, Chen, Shao, Guan, and Nie}]{shen2024mome}
Shen, L.; Chen, G.; Shao, R.; Guan, W.; and Nie, L. 2024.
\newblock MoME: Mixture of Multimodal Experts for Generalist Multimodal Large Language Models.
\newblock In \emph{Advances in Neural Information Processing Systems}, volume~37, 42048--42070.

\bibitem[{Song et~al.(2025)Song, Chen, Ding, Zhao, Zhao, Zhong, Ge, Ma, and Li}]{pdvla}
Song, W.; Chen, J.; Ding, P.; Zhao, H.; Zhao, W.; Zhong, Z.; Ge, Z.; Ma, J.; and Li, H. 2025.
\newblock Accelerating Vision-Language-Action Model Integrated with Action Chunking via Parallel Decoding.
\newblock \emph{arXiv preprint arXiv:2503.02310}.

\bibitem[{Unterthiner et~al.(2018)Unterthiner, Van~Steenkiste, Kurach, Marinier, Michalski, and Gelly}]{unterthiner2018towards}
Unterthiner, T.; Van~Steenkiste, S.; Kurach, K.; Marinier, R.; Michalski, M.; and Gelly, S. 2018.
\newblock Towards accurate generative models of video: A new metric \& challenges.
\newblock \emph{arXiv preprint arXiv:1812.01717}.

\bibitem[{Wu et~al.(2023)Wu, Jing, Cheang, Chen, Xu, Li, Liu, Li, and Kong}]{gr1}
Wu, H.; Jing, Y.; Cheang, C.; Chen, G.; Xu, J.; Li, X.; Liu, M.; Li, H.; and Kong, T. 2023.
\newblock Unleashing large-scale video generative pre-training for visual robot manipulation.
\newblock \emph{arXiv preprint arXiv:2312.13139}.

\bibitem[{Xie et~al.(2025)Xie, Shao, Chen, Zhou, Li, Liu, Zhang, and Nie}]{xie2025gui}
Xie, B.; Shao, R.; Chen, G.; Zhou, K.; Li, Y.; Liu, J.; Zhang, M.; and Nie, L. 2025.
\newblock GUI-explorer: Autonomous Exploration and Mining of Transition-aware Knowledge for GUI Agent.
\newblock In \emph{Annual Meeting of the Association for Computational Linguistics (ACL)}.

\bibitem[{Xu, Qiu, and She(2025)}]{xu2025vilp}
Xu, Z.; Qiu, Q.; and She, Y. 2025.
\newblock VILP: Imitation Learning with Latent Video Planning.
\newblock \emph{IEEE Robotics and Automation Letters}.

\bibitem[{Zhang et~al.(2025{\natexlab{a}})Zhang, Guo, Hu, Chen, Zhu, and Chen}]{zhang2025up}
Zhang, J.; Guo, Y.; Hu, Y.; Chen, X.; Zhu, X.; and Chen, J. 2025{\natexlab{a}}.
\newblock UP-VLA: A Unified Understanding and Prediction Model for Embodied Agent.
\newblock \emph{arXiv preprint arXiv:2501.18867}.

\bibitem[{Zhang et~al.(2025{\natexlab{b}})Zhang, Shao, Chen, Zhang, Zhou, Guan, and Nie}]{zhang2025falcon}
Zhang, R.; Shao, R.; Chen, G.; Zhang, M.; Zhou, K.; Guan, W.; and Nie, L. 2025{\natexlab{b}}.
\newblock Falcon: Resolving visual redundancy and fragmentation in high-resolution multimodal large language models via visual registers.
\newblock In \emph{Proceedings of the IEEE/CVF International Conference on Computer Vision}, 23530--23540.

\bibitem[{Zhao et~al.(2025{\natexlab{a}})Zhao, Lu, Kim, Fu, Zhang, Wu, Li, Ma, Han, Finn et~al.}]{cot}
Zhao, Q.; Lu, Y.; Kim, M.~J.; Fu, Z.; Zhang, Z.; Wu, Y.; Li, Z.; Ma, Q.; Han, S.; Finn, C.; et~al. 2025{\natexlab{a}}.
\newblock Cot-vla: Visual chain-of-thought reasoning for vision-language-action models.
\newblock \emph{arXiv preprint arXiv:2503.22020}.

\bibitem[{Zhao et~al.(2025{\natexlab{b}})Zhao, Lu, Kim, Fu, Zhang, Wu, Li, Ma, Han, Finn et~al.}]{cotvla}
Zhao, Q.; Lu, Y.; Kim, M.~J.; Fu, Z.; Zhang, Z.; Wu, Y.; Li, Z.; Ma, Q.; Han, S.; Finn, C.; et~al. 2025{\natexlab{b}}.
\newblock Cot-vla: Visual chain-of-thought reasoning for vision-language-action models.
\newblock In \emph{Proceedings of the Computer Vision and Pattern Recognition Conference}, 1702--1713.

\bibitem[{Zhao et~al.(2024)Zhao, Chen, Meng, Mao, Song, and Zhang}]{zhao2024vlmpc}
Zhao, W.; Chen, J.; Meng, Z.; Mao, D.; Song, R.; and Zhang, W. 2024.
\newblock Vlmpc: Vision-language model predictive control for robotic manipulation.
\newblock \emph{arXiv preprint arXiv:2407.09829}.

\bibitem[{Zhong et~al.(2025)Zhong, Huang, Li, Zhang, Liang, Yang, and Chen}]{dexgraspvla}
Zhong, Y.; Huang, X.; Li, R.; Zhang, C.; Liang, Y.; Yang, Y.; and Chen, Y. 2025.
\newblock DexGraspVLA: A Vision-Language-Action Framework Towards General Dexterous Grasping.
\newblock arXiv:2502.20900.

\bibitem[{Zhou et~al.(2025{\natexlab{a}})Zhou, Atreya, Tan, Pertsch, and Levine}]{autoeval}
Zhou, Z.; Atreya, P.; Tan, Y.~L.; Pertsch, K.; and Levine, S. 2025{\natexlab{a}}.
\newblock Autoeval: Autonomous evaluation of generalist robot manipulation policies in the real world.
\newblock \emph{arXiv preprint arXiv:2503.24278}.

\bibitem[{Zhou et~al.(2025{\natexlab{b}})Zhou, Zhu, Zhu, Wen, Liu, Xu, Meng, Cheng, Peng, Shen, and Feng}]{chatvla}
Zhou, Z.; Zhu, Y.; Zhu, M.; Wen, J.; Liu, N.; Xu, Z.; Meng, W.; Cheng, R.; Peng, Y.; Shen, C.; and Feng, F. 2025{\natexlab{b}}.
\newblock ChatVLA: Unified Multimodal Understanding and Robot Control with Vision-Language-Action Model.
\newblock arXiv:2502.14420.

\bibitem[{Zhu et~al.(2025{\natexlab{a}})Zhu, Yu, Feng, Burchfiel, Shah, and Gupta}]{zhu2025unified}
Zhu, C.; Yu, R.; Feng, S.; Burchfiel, B.; Shah, P.; and Gupta, A. 2025{\natexlab{a}}.
\newblock Unified world models: Coupling video and action diffusion for pretraining on large robotic datasets.
\newblock \emph{arXiv preprint arXiv:2504.02792}.

\bibitem[{Zhu et~al.(2025{\natexlab{b}})Zhu, Lyu, Yu, Shao, Zhou, and Nie}]{zhu2025emosym}
Zhu, Y.; Lyu, Y.; Yu, Z.; Shao, R.; Zhou, K.; and Nie, L. 2025{\natexlab{b}}.
\newblock EmoSym: A Symbiotic Framework for Unified Emotional Understanding and Generation via Latent Reasoning.
\newblock In \emph{Proceedings of the 33nd ACM International Conference on Multimedia}.

\bibitem[{Zhu et~al.(2025{\natexlab{c}})Zhu, Zhang, Yu, Shao, Tan, and Nie}]{zhu2025uniemo}
Zhu, Y.; Zhang, L.; Yu, Z.; Shao, R.; Tan, T.; and Nie, L. 2025{\natexlab{c}}.
\newblock UniEmo: Unifying Emotional Understanding and Generation with Learnable Expert Queries.
\newblock \emph{arXiv preprint arXiv:2507.23372}.

\end{thebibliography}
\cleardoublepage
\appendix
We provide comprehensive supplementary material to support the methodology, implementation, and analysis introduced in the main paper on H-GAR. 

\section*{A Implementation Details}

\subsection{A.1 Training Details.}
\paragraph{Simulation Training Setup.} We train our model on three simulation benchmarks: PushT, PushT-M and Libero-10. Each benchmark comprises 16 observation steps and 8 action steps, with a prediction horizon of 32. By default, we enable mixed-precision training (FP16) and apply an exponential moving average (EMA) to enhance training efficiency and stability.

PushT and PushT-M use RGB observations with a resolution of 96×96 and a 2-dimensional action space. Libero-10 use 128×128 RGB observations and a 10-dimensional action space encoded with 6D rotation representation. A batch size of 64 is used across all benchmarks. We train using the Adam optimizer with a cosine learning rate schedule and a 1000-step linear warmup. The best-performing checkpoint is selected based on validation performance.

\paragraph{Real-World Training Setup.} In real-world experiments, we use a batch size of 32 and train the model with mixed-precision bfloat16. Training is conducted using the Adam optimizer with a cosine learning rate schedule and a 1000-step linear warmup. Exponential moving average (EMA) is applied throughout training. We evaluate checkpoints after each epoch based on the L2 distance between predicted and ground-truth actions on the validation set, retain the top 5 models, and select the best-performing checkpoint for final evaluation.


\subsection{A.2 More Network Details.}

Early autoregressive visual modeling approaches, such as VQGAN-based methods~\cite{esser2021VQGAN}, convert images into discrete token sequences via vector quantization. During training, a subset of tokens is randomly masked, and the model learns to reconstruct the missing content. This setup enables autoregressive generation during inference, where images or videos are synthesized step-by-step by progressively unmasking tokens. While effective, this method often suffers from quantization artifacts and loss of fine-grained details, resulting in blurry or distorted outputs.

To mitigate this limitation, more recent work has shifted towards continuous latent modeling. Instead of predicting discrete indices, these methods directly model continuous latent features using a diffusion process, which allows for richer and smoother reconstructions. For example, Recent work~\cite{li2024autoregressive} replaces discrete token modeling with a continuous diffusion-based representation, achieving improved visual quality.

Our H-GAR model extends the continuous autoregressive framework with two modules that support fine-grained, goal-consistent prediction through explicit observation-action interaction. Given a coarse action sequence and predicted goal observation from a base diffusion model, the Goal-Conditioned Observation Synthesizer (GOS) generates intermediate visual frames that reflect expected motion. These are used by the Interaction-Aware Action Refiner (IAAR) to refine the coarse actions by integrating visual feedback and temporal cues from a Historical Action Memory Bank. This bidirectional interplay allows the model to align action and observation generation in a coarse-to-fine manner, promoting coherence over long horizons.

\begin{figure*}[t]
\centering
\includegraphics[width=1.0\textwidth]{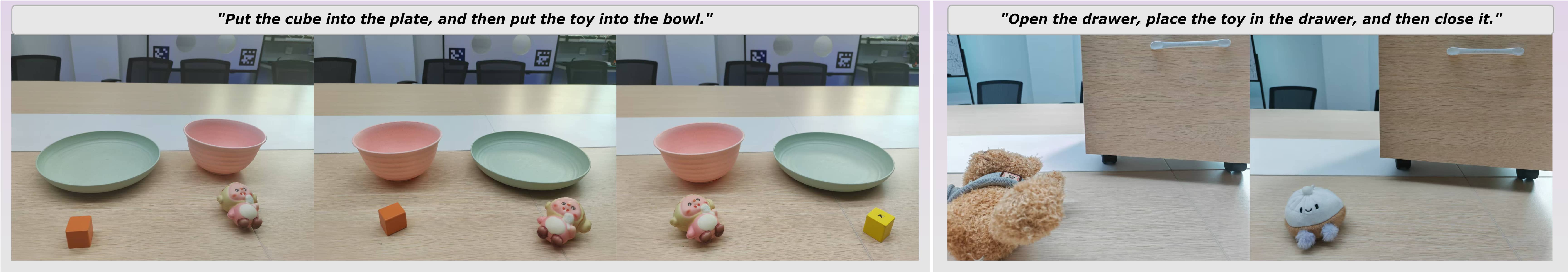}
\caption{\textbf{Initial States for Data-Augmented Tasks} Representative initial observations two relatively complex tasks(Task 1 and Task 2), where moderate data augmentation is applied. Object attributes (e.g., color, size) and spatial layouts are systematically varied to introduce controlled diversity. This design assesses model robustness and enhances cross-instance generalization while maintaining intra-task consistency.} 
\label{vis_dataaug}
\end{figure*}
\section*{B Experimental Details}
\subsection*{B.1 Simulation Benchmarks}

\paragraph{Single-Task Evaluation.} In this setting, we conduct experiments on the \textbf{PushT} task, which requires the agent to push a gray “T” to align it with the target “T” located at the center of the scene. 

\paragraph{Multi-Task Evaluation.} In the multi-task evaluation setting, a single policy is trained to handle diverse task goals specified by either image or language inputs.

\begin{itemize}
    \item \textbf{PushT-M}: The original PushT task is modified by randomizing the position of the target “T” across episodes, introducing spatial variability and increasing task complexity. This setting requires the agent to generate flexible plans and adapt its behavior accordingly.

    \item \textbf{Libero-10}: Libero-10 (also referred to as Libero-Long) is a suite of long-horizon robotic manipulation tasks involving diverse objects, spatial configurations, and complex goal descriptions. Each task requires executing a sequence of temporally extended actions conditioned on high-level language instructions. For instance, the task “put the yellow and white mug in the microwave and close it” demands multi-step reasoning, goal grounding, and precise spatial interaction. These tasks are designed to evaluate an agent’s ability to integrate visual and language inputs, maintain temporal coherence, and generalize across varying object arrangements and interaction patterns.
    
\end{itemize}

\subsection*{B.2 Real-World Setup}
To evaluate the real-world applicability of H-GAR, we deploy it on the Cobot Agilex ALOHA manipulation platform. The evaluation includes four diverse tasks, each involving only single-arm manipulation. On two relatively complex tasks, moderate data augmentation is applied by varying object attributes and rearranging spatial layouts to assess the model's robustness and generalization capabilities.
\paragraph{Task Descriptions.}
Below are the instructions and descriptions for Tasks 1–4:
\begin{itemize}
    \item \textbf{Task 1}: ``\textit{Put the cube into the plate, and then put the toy into the bowl.}''\\
    A sequential two-object manipulation task requiring temporal coordination and object discrimination. This operation involves two consecutive subtasks executed with a single robotic arm: 1)``\textit{Place the cube into the plate}'', followed immediately by 2) ``\textit{Put the toy into the bowl}''. Task success demands both subtasks to be completed in strict chronological order using the same arm without error. Performance is evaluated via subtask-specific success rates. We present the success rates corresponding to each subtask.
    \item \textbf{Task 2}: ``\textit{Open the drawer, place the toy into the drawer, and then close it.}''\\
    A sequential three-step manipulation task requiring spatial reasoning. This operation involves three consecutive subtasks executed with a single robotic arm: 1) ``\textit{Open the drawer}'', 2) ``\textit{Place the toy into the drawer}'', and 3) ``\textit{Close the drawer}''. Task success requires all subtasks to be completed in chronological order using the same arm, with the drawer maintaining structural integrity. Performance is evaluated through subtask success rates. We present the success rates corresponding to each subtask.
    \item \textbf{Task 3}: ``\textit{Grasp the left edge of the towel and move it to the right, folding it in half.}''\\
    A deformable fabric manipulation task evaluating single-fold execution on pliable surfaces. This operation requires grasping the towel and performing one continuous folding action to achieve a stable bent configuration. Task success is determined solely by the final folded state maintaining structural integrity without slippage. Performance is reported through overall success rate.
    \item \textbf{Task 4}: \textit{``Pick up the mouse and place it on the mouse pad.''}\\ 
    A precision manipulation task requiring targeted grasping and planar placement. This operation involves transferring the mouse from its initial position to the designated pad surface in a single continuous motion. Task success is determined exclusively by the mouse resting stably within the pad boundaries without displacement. Performance is reported through the overall success rate.
\end{itemize}

\begin{figure*}[t]
\centering
\includegraphics[width=1.0\textwidth]{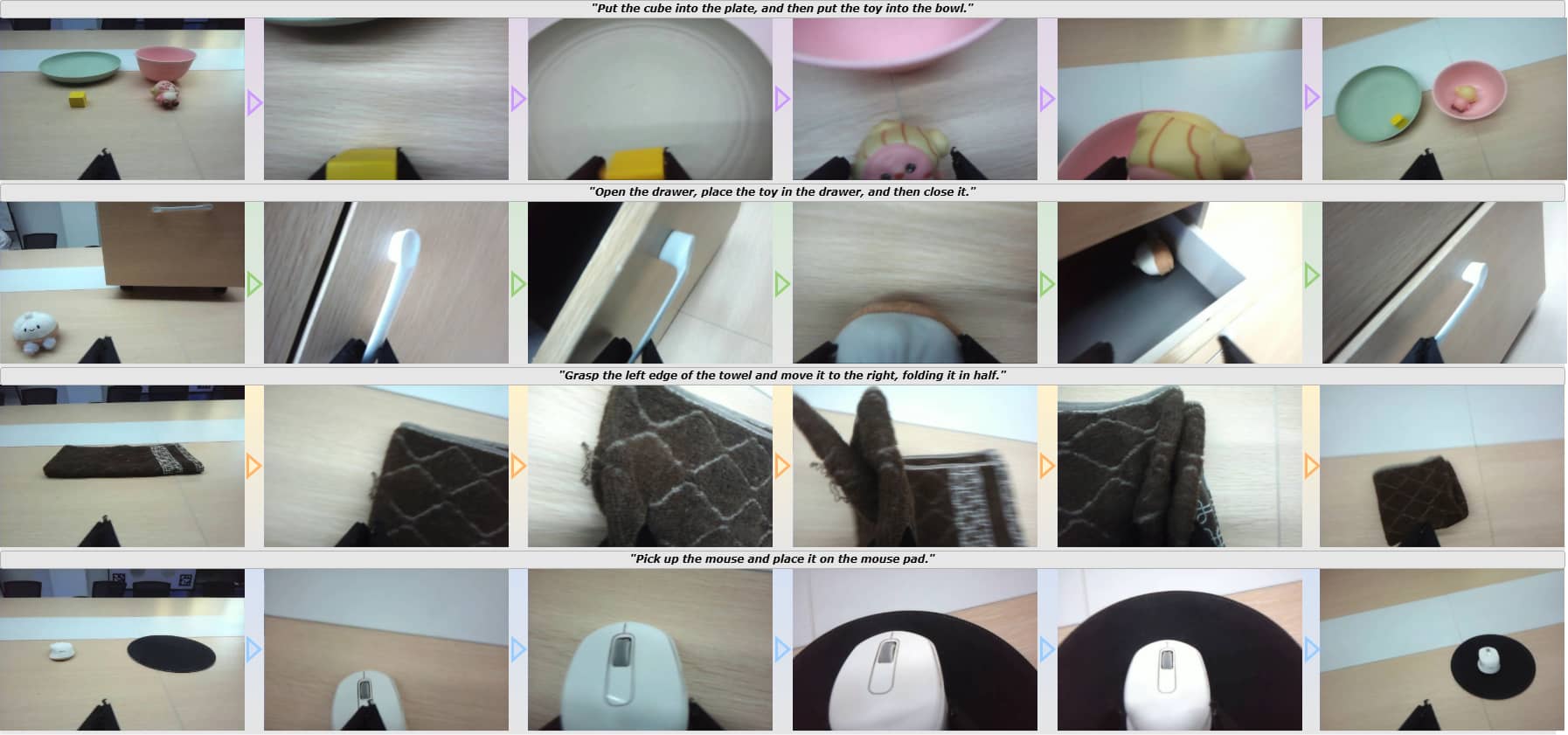}
\caption{\textbf{Real-world Manipulation Workflows for Tasks 1-4} This figure illustrates H-GAR's execution of Tasks 1-4 through \textit{Left Wrist Camera} observations, capturing initial setups and step-by-step workflows. These workflows exemplify H-GAR's temporal reasoning for step coordination and adaptation to task complexity. Full video demonstrations are available in supplementary materials.}
\label{vis_firstperson}
\end{figure*}

\paragraph{Data Collection and Augmentation}
Real-world training data for the Cobot Agilex ALOHA robot is collected via human teleoperation. To enhance generalization, moderate data augmentation—varying initial object poses, object attributes, and tabletop layouts—is applied exclusively to Tasks 1 and 2, the initial observations for each group in Tasks 1–2 are illustrated in Figure~\ref{vis_dataaug}. Details of the data collection and augmentation procedures for each task are provided below.
\begin{itemize}
    \item \textbf{Task 1}: 45 expert demonstrations were collected and divided into three equal groups (15 per group). Variations across groups included cube properties, plate-bowl positioning, and overall object arrangements.
    \item \textbf{Task 2}: We acquired 45 expert demonstrations, partitioned into two distinct groups (25 and 20 demonstrations, respectively). These groups featured variations in toy categories, sizes, and colors.
    \item \textbf{Task 3}: 30 expert demonstrations were recorded without augmentation, preserving the original teleoperation scenarios.
    \item \textbf{Task 4}: Similarly, 30 unaugmented demonstrations were collected, maintaining the data in its original teleoperated state.
\end{itemize}

\begin{figure*}[t]
\centering
\includegraphics[width=0.92\textwidth]{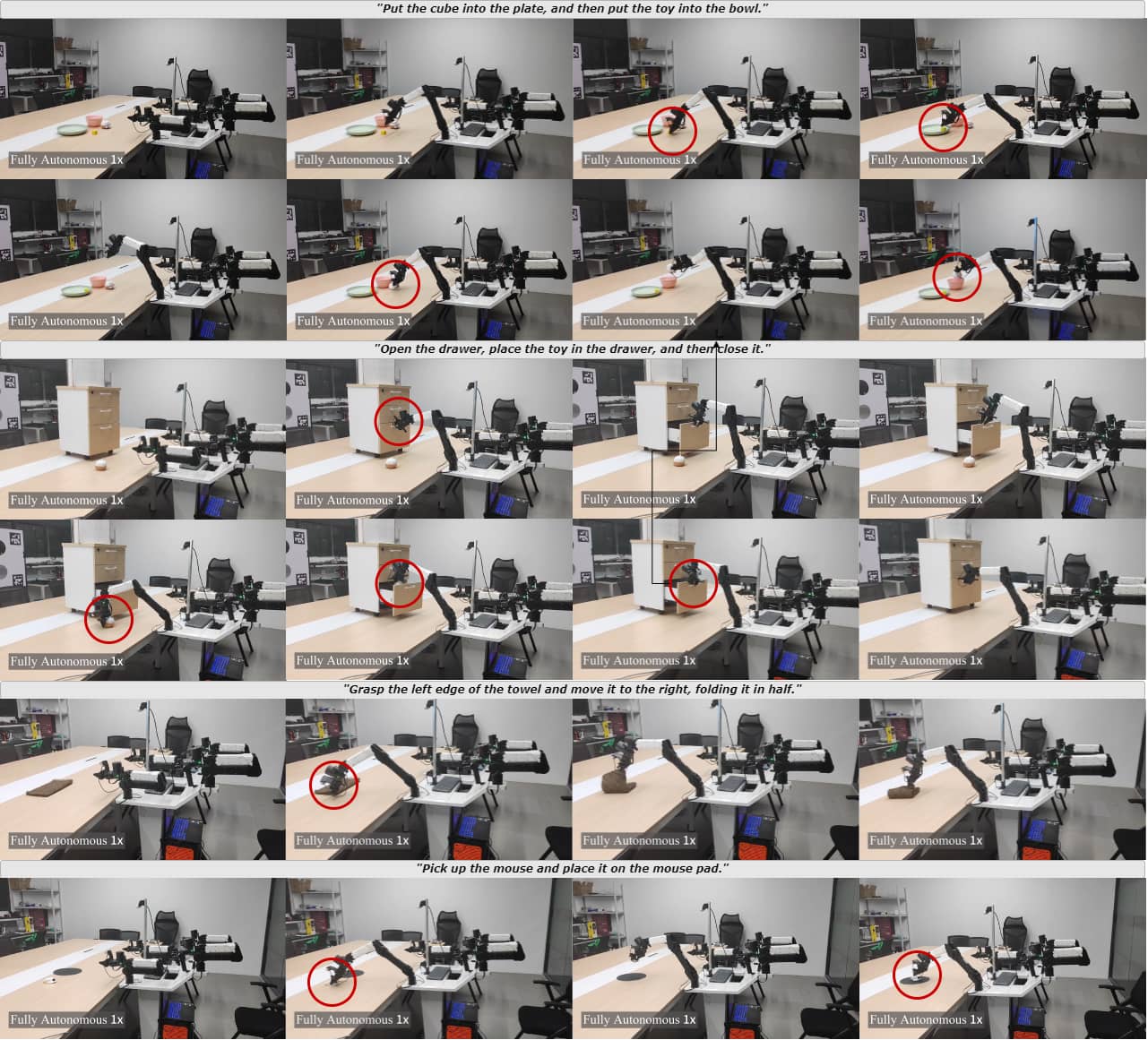}
\caption{\textbf{Third-person visualization depicting H-GAR executing a manipulation task.}The corresponding video is included in the supplementary materials. The details of the gripper are marked with red circles for emphasis.}
\label{vis_thirdperson}
\end{figure*}

\subsection*{B.3 Baselines}
We compare H-GAR with a diverse set of strong baselines encompassing both action generation and observation prediction paradigms. These methods represent the current state of the art in visuomotor policy learning and generative modeling for embodied agents.

\begin{itemize}
    \item \textbf{Diffusion Policy-C/T~\cite{Chi2023diffusion_policy}} are two variants of Diffusion Policy, where the “C” version adopts a CNN-based policy backbone, while the “T” version replaces it with a Transformer-based architecture. Both variants generate fine-grained actions based on past observations and are widely adopted in robotic imitation learning benchmarks.
    \item \textbf{OpenVLA~\cite{Kim2024openvla}} is a vision-language-action model built on a pretrained LLaMA-2 backbone. It supports both image and language modalities and is designed to enable multi-task generalization.
    \item \textbf{SpatialVLA~\cite{spatialvla}} enhances vision-language-action modeling by incorporating egocentric 3D spatial reasoning. It introduces Ego3D Position Encoding, which integrates depth cues into visual features without needing camera calibration, and Adaptive Action Grids, which discretize actions into spatial tokens. 
    \item \textbf{CoT-VLA~\cite{cotvla}} introduces visual chain-of-thought reasoning by first generating a subgoal image based on the instruction and current observation, then predicting actions to reach it. This two-stage process improves planning and generalization in complex tasks. CoT-VLA is built on VILA-U with a hybrid attention mechanism. It leverages large-scale pretraining on both action-labeled and unlabeled videos.
    \item \textbf{UniPi~\cite{unipi}} formulates policy learning as text-conditioned video generation. It first synthesizes future observations based on the instruction, then infers actions via an inverse dynamics model. This decoupled formulation promotes generalization and adaptability across tasks.
    \item \textbf{UVA~\cite{li2025unified}} is a generative model for embodied agents that produces a unified latent representation, which is decoded via separate diffusion heads for visual trajectory and action sequence generation. This design supports joint modeling while maintaining modularity between visual and control pathways.
    \item \textbf{VQ-BeT~\cite{vqbet}} employs a single-layer MLP encoder with 128 dimensions and a residual vector quantization module containing roughly 1024 codes. The model processes sequences using an observation window of 10 timesteps and an action window size of $T=5$, as longer sequences (e.g., $T=32$) may lead to information degradation due to over-compression.
    \item \textbf{QueST~\cite{quest}} employs a 6-layer transformer decoder with 6 attention heads and an embedding dimension of 384. It generates actions using a discrete vocabulary of size 1000, with token embeddings matching the model dimension. During inference, beam search is performed with a width of 5 and temperature set to 1.0, predicting over action blocks of 8 timesteps.
    \item \textbf{STAR~\cite{star}} is a skill-based control framework that incorporates RaRSQ to preserve geometric structure during skill quantization and a Causal Skill Transformer (CST) to autoregressively model skill dependencies and offsets. It shows strong performance on long-horizon manipulation tasks.
    \item \textbf{PD-VLA~\cite{pdvla}} is a vision-language-action model designed to improve decoding efficiency when integrated with action chunking. Traditional autoregressive VLA models suffer from inference latency due to sequential action token generation. PD-VLA addresses this by reformulating the decoding process into a parallel fixed-point iteration system, enabling all action tokens to be predicted simultaneously in each iteration. This approach accelerates inference without requiring retraining or architectural modifications.
\end{itemize}
\section*{C Supplementary Qualitative Analysis}
\paragraph{Real-World Visualization.}  
Across diverse real-world tasks with varying instructions, H-GAR demonstrates sequential task execution and spatial reasoning, accurately interpreting long-horizon commands and generating coherent, chronologically ordered action sequences. These examples further highlight the model's core capabilities: maintaining cross-modal consistency essential for precise object recognition and discrimination, and performing complex temporal reasoning for step coordination. Figure~\ref{vis_firstperson} illustrates the real-world manipulation workflows for Tasks 1-4, utilizing \textit{Left Wrist Camera} observations for all tasks. Figure~\ref{vis_thirdperson} shows a third-person view demonstration of H-GAR performing one manipulation task, while videos for all four tasks are available in the supplementary materials.


\end{document}